\documentclass{article} 
\usepackage[final]{colm2026_conference}

\usepackage{microtype}
\usepackage{hyperref}
\usepackage{url}
\usepackage{booktabs}
\usepackage{multirow}
\usepackage{amsmath,amssymb}
\usepackage{algorithm}
\usepackage{algpseudocode}
\usepackage{enumitem}

\usepackage{graphicx}
\usepackage{tablefootnote}
\usepackage{wrapfig}

\usepackage{listings}
\usepackage{xcolor}

\lstdefinestyle{prompt}{
  basicstyle=\small\ttfamily,
  breaklines=true,
  breakatwhitespace=false,
  frame=single,
  rulecolor=\color{gray!40},
  backgroundcolor=\color{gray!5},
  aboveskip=6pt,
  belowskip=6pt,
  xleftmargin=4pt,
  xrightmargin=4pt,
  columns=fullflexible,
  literate={"}{\texttt{\char34}}1,
}
\lstdefinestyle{lesson}{
  basicstyle=\small\ttfamily,
  breaklines=true,
  breakatwhitespace=false,
  frame=single,
  rulecolor=\color{blue!20},
  backgroundcolor=\color{blue!3},
  aboveskip=6pt,
  belowskip=6pt,
  xleftmargin=4pt,
  xrightmargin=4pt,
  columns=fullflexible,
  literate={"}{\texttt{\char34}}1,
}

\usepackage[most]{tcolorbox}

\definecolor{boxblue}{RGB}{235,247,255}
\definecolor{borderblue}{RGB}{145,195,230}

\newtcolorbox{titleabstractbox}{
  enhanced,
  colback=boxblue,         
  colframe=borderblue,      
  boxrule=0.8pt,            
  arc=4mm,                  
  left=6mm,
  right=6mm,
  top=5mm,
  bottom=6mm,
  width=\textwidth
}

\usepackage{xcolor}

\usepackage{lineno}

\definecolor{darkblue}{rgb}{0, 0, 0.5}
\hypersetup{colorlinks=true, citecolor=darkblue, linkcolor=darkblue, urlcolor=darkblue}

\title{
The Optimizer Is the Agent: Reasoning-Driven Search across Prompts, Programs, and ML Workflows
}

\newcommand{\samethanks}{\footnotemark[2]}

\author{
Junbo Li$^{1}$\thanks{Work done during internship at Snowflake.}
\And
Boyi Liu$^{2}$
\And
Canwen Xu$^{2}$
\And
Yite Wang$^{2}$
\AND
Yuxiong He$^{2}$
\And
Zhangyang Wang$^{1}$\thanks{These authors contributed equally and share last authorship.}
\And
Qiang Liu$^{1}$\samethanks
\And
Zhewei Yao$^{2}$\samethanks
\AND
{} \And
{\normalfont $^1$ The University of Texas at Austin \quad\quad $^2$ Snowflake}
\And {}
}

\begin{document}

\maketitle

\begin{abstract}
Recent systems for optimizing prompts, programs, and ML workflows typically rely on explicit outer-loop controllers such as evolutionary search, bandits, or textual-gradient methods. We ask a fundamentally different question: how much of this search policy can be internalized by a single tool-using agent? We present \textbf{ReASearch}, a unified framework for reasoning-driven optimization in which the agent autonomously decides what to evaluate, how to diagnose failures, which edits to make, and when to verify or restart. Rather than serving only as a proposal generator guided by hand-designed heuristics, the agent actively analyzes outcomes, allocates budget, and refines its strategy over long horizons through persistent memory. With a shared agent loop and domain-specific tools, ReASearch instantiates the exact same scaffold to optimize prompts, programs, and ML workflows. Across 14 diverse tasks, it is competitive with and mostly better than specialized optimization systems, achieving gains of 2\% to 40\% over strong domain-specific baselines, and discovering solutions that improve on prior human best-known results in Circle Packing. Crucially, we observe that complex search behaviors, which are typically implemented by explicit controllers, emerge naturally from the agent's reasoning process. \footnote{Codes are available at \url{https://github.com/Snowflake-AI-Research/ReASearch}}
\end{abstract}

\section{Introduction}

Optimizing text-based artifacts, such as system prompts, programs, and training configurations, is a central challenge in modern AI.
Traditional optimization methods often struggle in these settings because they operate over large, discrete spaces while lacking access to the rich semantic structure of natural language and code. This limitation has motivated a growing body of work that uses large language models as semantic optimizers: LLMs propose structured edits based on execution feedback, while external search mechanisms such as Bayesian optimization \citep{kang2025textbo,mipro,schneider2024hyperband,bergstra2011algorithms}, bandits \citep{cemri2026adaevolve,wu2025llm,shi2024efficient,li2026efficient,pryzant2023automatic}, and evolutionary search \citep{secheresse2025gaapo,guo2023evoprompt,fernando2023promptbreeder} guide exploration and selection. 

This line of research has advanced rapidly, encompassing controllers that range from prompting-based methods that optimize against score histories \citep{opro}, to textual-gradient frameworks \citep{khattab2023dspy,yuksekgonul2024textgrad,adalflow}, to planning- and evolution-based systems that maintain candidate sets and search over revisions \citep{wang2023promptagent, agrawal2025gepa, novikov2025alphaevolve, cemri2026adaevolve}, as well as LLM-generated adaptive controllers for meta-heuristics \cite{zelikman2023self,van2024llamea,shi2026generalizable}.
Yet across this diversity, a central design choice remains largely unchallenged: most strategic search decisions still live outside the model. The external algorithmic controller decides which candidate to branch from, which evaluations to run, how to allocate budget, and how to recover from stagnation; the LLM merely contributes semantic edits within that hard-coded loop. This leaves open a more fundamental question than whether LLMs can optimize at all: \emph{how much of the search policy can be internalized by a single reasoning agent?}

\begin{figure}
    \centering
    \includegraphics[width=0.85\linewidth]{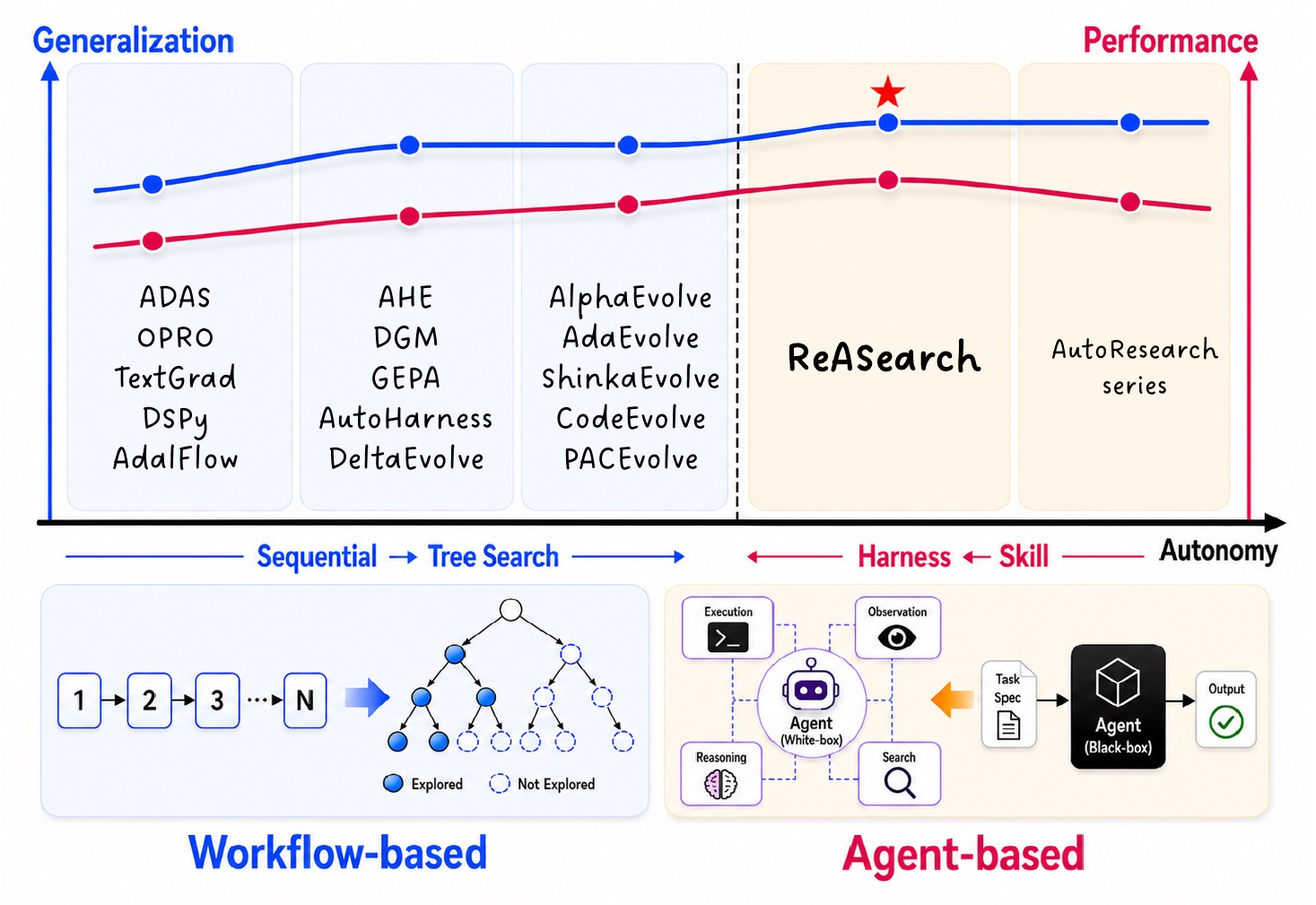}
    \vspace{-6mm}
    \caption{Positioning ReASearch relative to existing methods.}
    \vspace{-1mm}
    \label{fig:position}
\end{figure}

We study this question through Reasoning-driven Agentic Search (ReASearch), a controller-light framework that places a tool-using LLM agent fully in charge of the optimization loop. While recent proof-of-concept projects have demonstrated agents automating specific ML pipelines, ReASearch provides a systematic evaluation of internalized search across diverse modalities. The agent is equipped with domain-specific tools for evaluation, analysis, editing, and memory across three distinct settings: prompt optimization, program evolution, and ML workflow optimization. There is no fixed outer-loop search procedure: core optimization operations are all packaged as tools. The agent decides what to test next, when to verify, when to revert, and when to stop exploring to exploit a stronger candidate. 
We provide a concise comparison of ReASearch with existing methods in Figure~\ref{fig:position}.

Across a benchmark suite spanning 14 tasks, the same agent scaffold is highly competitive with specialized systems built around explicit search procedures. Furthermore, its trajectories exhibit recurring optimizer-like behaviors that are not hard-coded in the controller: double-verification of promising gains, reuse of similar past failures, deliberate reverting after unproductive branches, and adaptive exploration. These results suggest that, given rich-feedback environments, frontier tool-using agents can internalize a substantial fraction of the search logic previously delegated to external mathematical meta-heuristics.

\paragraph{Contributions.} Our contributions can be summarized as follows:
\begin{itemize}[nosep,leftmargin=*]
\item We cast text-based optimization as a sequential reasoning problem and present ReASearch, a unified, controller-light agent framework for optimizing prompts, programs, and ML workflows using a single shared agent design.
\item Across three distinct domains and 14 diverse tasks, ReASearch consistently matches or outperforms strong, domain-specific baselines (by 2\% to 40\%) within the same budgets, and in some cases discovers solutions surpassing prior human best-known results.
\item We conduct a systematic analysis of agent trajectories and identify recurring reasoning patterns that emerge during search, including candidate verification, hypothesis-driven revision, reuse of prior lessons, revert-based recovery, and adaptive exploration under budget constraints. These behaviors arise without any additional heuristic control.
\end{itemize}

\subsection{Related Work}
ReASearch relates to LLM-based text optimization, autonomous ML agents, and code agents; across these, the recurring pattern is an external controller that owns the search policy while the LLM contributes only local edits. We defer a detailed discussion to Appendix~\ref{app:related}.

\section{Method}
\label{sec:method}

We formulate the objective (\S\ref{sec:formulation}), present the general agent architecture (\S\ref{sec:architecture}), and describe domain-specific tool designs for 3 task categories (\S\ref{sec:domains}) with details in Appendix \ref{app:detail}.

\begin{figure}
    \centering
    \vspace{-7mm}
    \includegraphics[width=0.8\linewidth]{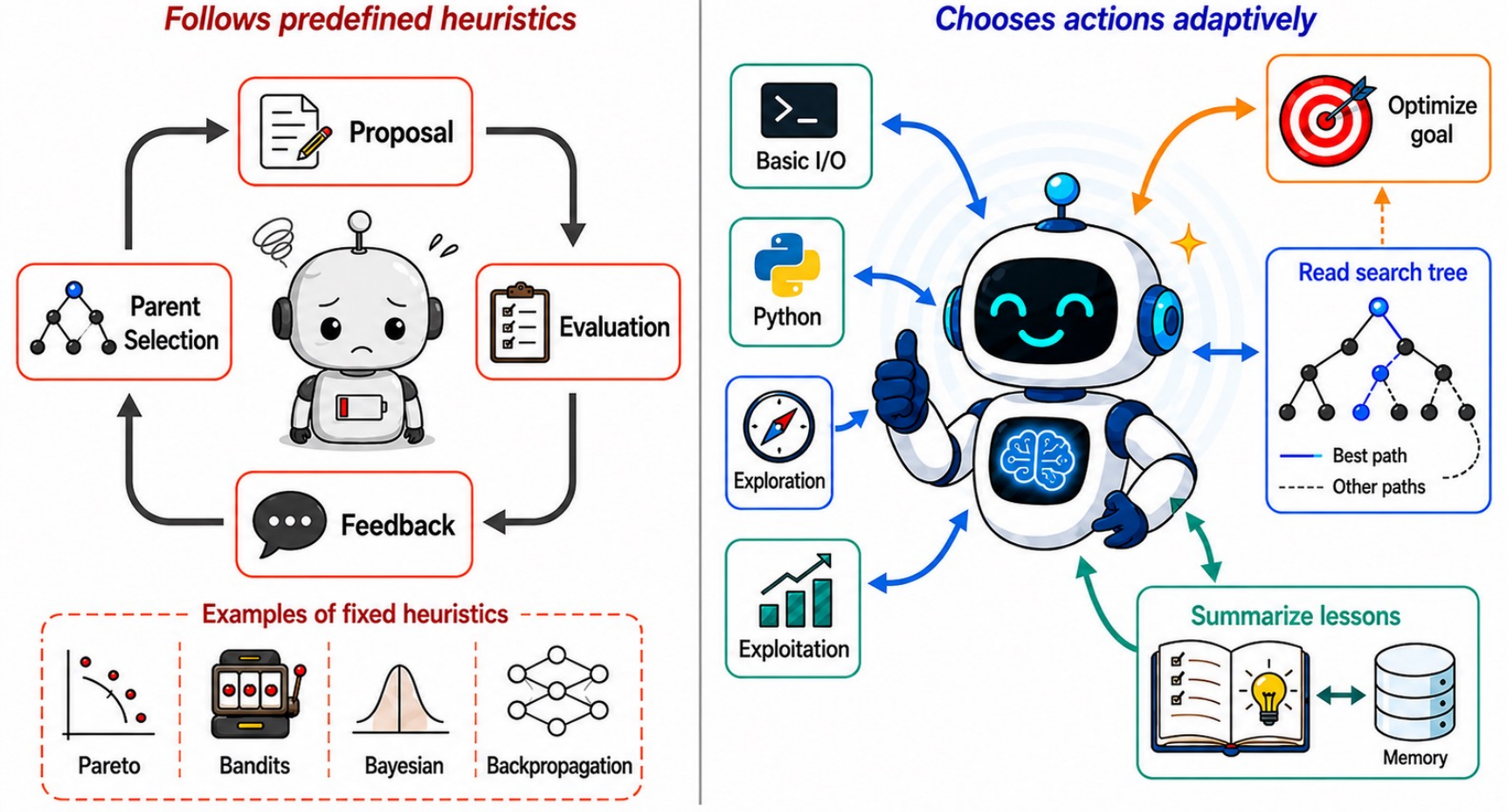}
    \vspace{-3mm}
    \caption{Prior methods use LLMs primarily as mutators, with candidate selection and optimization goals determined by external heuristic algorithms. In contrast, ReASearch exposes the entire optimization process through tools, giving the agent full control over the search procedure and enabling it to solve a wide range of tasks.}
    \label{fig:main}
\end{figure}

\subsection{Problem formulation}
\label{sec:formulation}

Let \(\mathcal{Z}\) denote the space of textual artifacts, such as prompts, programs, and training configurations. Given a task instance \(x \sim \mathcal{D}\) and an artifact \(z \in \mathcal{Z}\), the system generates an output \(Y \sim P(\cdot \mid x, z)\) with reward \(r(x, Y)\). Our objective is to find
\begin{equation}
z^\star \in \arg\max_{z \in \mathcal{Z}}
\;\mathbb{E}_{x \sim \mathcal{D}}
\left[
\mathbb{E}_{Y \sim P(\cdot \mid x, z)}[r(x, Y)]
\right].
\label{eq:obj}
\end{equation}
This formulation is general and naturally simplifies to task-free optimization as a special case. In scenarios where the objective is independent of specific input instances $x$, such as in algorithmic discovery or geometric optimization tasks, the outer expectation over $\mathcal{D}$ vanishes. Consequently, the goal reduces to maximizing the expected reward of the artifact $z$ itself. The key challenge is how to search $\mathcal{Z}$. Prior work typically uses external meta-heuristics, such as evolutionary algorithms, bandits, or textual gradient methods. We ask whether an LLM agent with appropriate tools can instead perform this search through reasoning alone.

\subsection{General agent architecture}
\label{sec:architecture}

We implement a simple code agent with basic file I/O, Python/Bash execution, and a lightweight memory module. At each turn, the agent receives a \emph{system prompt} encoding domain guidance and the current optimization state, along with its \emph{interaction history}, and reasons about which tool to invoke next. The core loop itself is task-agnostic: all task-specific information is provided through two interfaces only, the tool set \(\mathcal{A}_{\mathcal{T}}\) and the task-specific \emph{system prompt}. As a result, the same agent can optimize prompts, programs, and training scripts by changing only the available tools and the prompt.
Among all these tools, \texttt{python\_exec} is the most important: it transforms the agent from a pure text reasoner into a \textit{computational} reasoner that can write analysis scripts, compute statistics over evaluation logs, and programmatically identify failure patterns. 
Figure \ref{fig:main} visualizes our framework and compares it with prior methods.

\paragraph{Context management and memory}
To support long-horizon optimization, we equip the agent with both context compression and persistent memory. When token usage exceeds a threshold, or when the agent chooses to prune its context, it invokes a compaction tool to summarize the interaction history. The agent also maintains a persistent \texttt{lessons.md} file and is periodically prompted to read from or update it. This file records the most important lessons from the run, including what worked, what failed, and what to try next, along with key empirical observations, such as which methods achieved which levels of performance under which conditions.
\subsection{Domain-specific tool design}
\label{sec:domains}

\subsubsection{Prompt optimization.}

\begin{figure}
    \centering
    \vspace{-5mm}
    \includegraphics[width=0.9\linewidth]{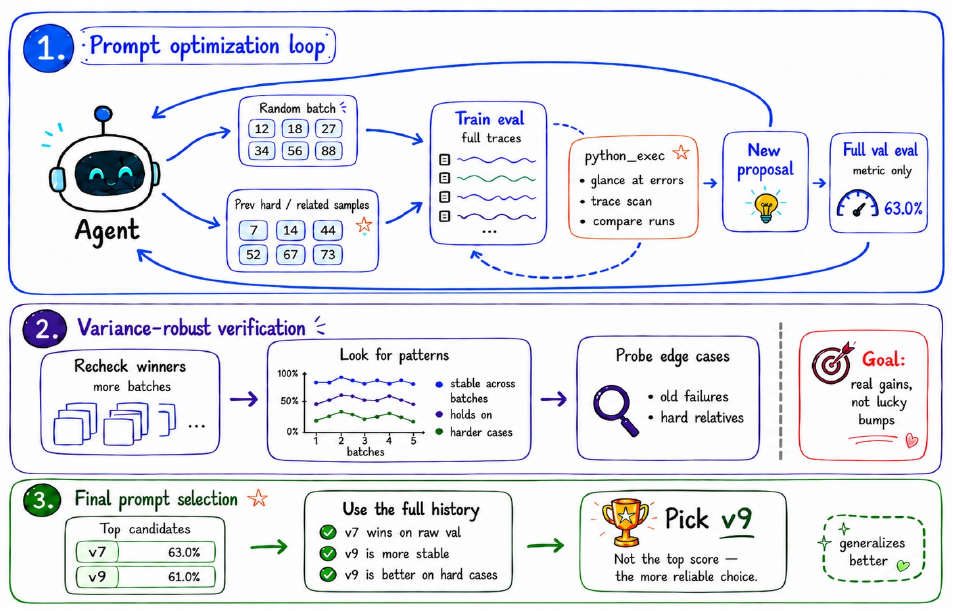}
    \vspace{-3mm}
    \caption{We visualize an example agent trajectory for prompt optimization, highlighting automatic verification and final prompt selection. None of this process is hardcoded; it emerges from the guiding instruction, with agents freely deciding each next step.}
    \label{fig:prompt}
\end{figure}

\paragraph{Tools}
The agent optimizes a system prompt for a student LLM on a dataset with train, validation, and test splits. Unlike prior hard-coded pipelines that repeatedly sample random training minibatches, revise the prompt, and validate new proposals, 
we expose training and validation evaluation as tools that the agent can invoke adaptively based on its reasoning and remaining budget.
First, \texttt{get\_next\_minibatch} samples the next minibatch of $B$ training examples, where $B$ is chosen by the agent according to its current budget. Second, \texttt{call\_student\_model\_batch} evaluates a prompt candidate on a specified batch of example indices, which may correspond either to newly sampled training examples or to previously identified difficult cases. The agent has access to the full evaluation trajectories on these training examples. Third, \texttt{validate\_candidate} evaluates a prompt candidate on the validation set and returns only aggregate accuracy, without per-example trajectories, to reduce overfitting to validation feedback. In addition, the agent can call the general-purpose \texttt{python\_exec} tool to inspect prior evaluation results, compute summary statistics, and perform more detailed diagnosis.

\paragraph{Variance-robust effective verification}
The full autonomy is especially important because prompt evaluation is often high variance, making purely metric-based control brittle. Figure~\ref{fig:prompt} illustrates a representative agent trajectory for prompt optimization. With access to the tools, the agent naturally adapts its search strategy. For example, we observe the agent rechecking promising prompt candidates on additional training samples before validation, analyzing whether gains reflect genuine capability improvements or superficial pattern matching, examining score trajectories across diverse examples to assess robustness, and deliberately probing edge cases corresponding to earlier failure modes. Behaviors that would previously require hand-designed algorithmic components instead emerge directly from reasoning over the evaluation history.

\paragraph{Final prompt selection}
The final prompt is not determined solely by the candidate with the best validation metric, as is typical in prior work. Instead, after the optimization run, the final agent turn is instructed to select the final prompt, or synthesize a new one, by reasoning over the full evaluation history and the lessons accumulated throughout the run. This is especially useful in data-scarce settings such as AIME and Terminal-Bench, where validation feedback can be highly variable. In such cases, we observe that the agent's final choice can outperform the prompt with the highest validation score on the test set.

\subsubsection{Program evolution.}

\begin{figure}
    \centering
    \vspace{-4mm}
    \includegraphics[width=0.72\linewidth]{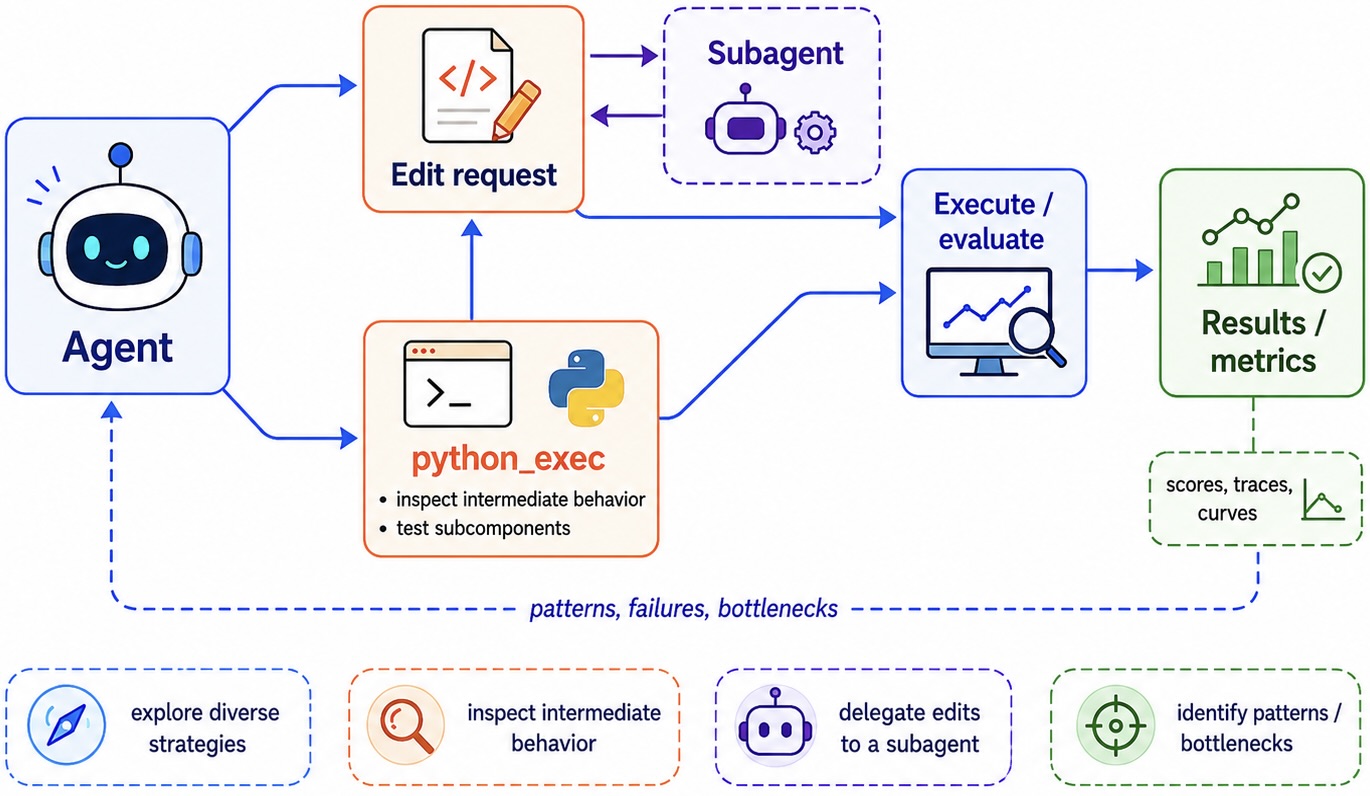}
    \vspace{-1mm}
    \caption{Program evolution and ML workflow optimization exhibit similar patterns: the agent actively uses Python for reasoning and small-scale experiments before running expensive evaluations, while delegating targeted code edits to a separate agent to keep the main agent’s context clean.}
    \label{fig:code}
\end{figure}

The agent optimizes a program for a single hard problem, such as a challenging optimization task or a pattern induction problem in ARC-AGI-2 \citep{chollet2025arc}, where fundamentally different algorithmic strategies may all be plausible. This setting requires the agent to explore diverse approaches, reason about why one strategy outperforms another on particular problem structures, and accumulate insights across attempts. These requirements align naturally with the strengths of our framework.

\paragraph{Tools}
As in prompt optimization, program evaluation is exposed as a tool, \texttt{evaluate}, which the agent can invoke whenever it deems appropriate. At the start of the run, the agent reads the baseline program. Thereafter, whenever it decides to modify the program, it calls the \texttt{edit\_code} tool, which routes the requested changes to a separate subagent that applies the edit while keeping the main agent context clean. We additionally use a verification subagent that checks whether each edit is valid and respects task constraints. Higher-level analysis and diagnosis are deliberately kept inside the main agent: this preserves reasoning context across turns (which can be compacted into useful memory), avoids information loss between agents, and reduces coordination complexity. Figure~\ref{fig:code} illustrates this pattern, which program evolution and ML workflow optimization share.

\paragraph{Lesson accumulation and sub-program-based analysis}
Compared with prior evolutionary systems, our framework offers two main advantages. First, the cycle of repeated attempts and lesson accumulation enables the agent to improve continuously on difficult tasks. Rather than treating each mutation as an isolated step, the agent can form and refine hypotheses over time by integrating evidence across many trials. Second, the agent can make extensive use of the general-purpose \texttt{python\_exec} tool to inspect intermediate behavior, analyze outputs, test subcomponents in isolation, and reason directly about the structure of the problem. This allows it to move beyond shallow mutation and perform deliberate diagnosis, verification, and refinement. 

\subsubsection{ML workflow optimization.}
This category covers machine learning training tasks in which improvement may come from many sources, including hyperparameter tuning, feature engineering, architectural changes, loss design, optimizer choice, and regularization. The agent reads the training code, runs limited-duration experiments via \texttt{run\_experiment}, observes the resulting metrics and training dynamics, and proposes targeted modifications to address the current bottlenecks. The overall tool design is similar to that in program evolution, but the instructions and execution pipeline are specialized for ML workflows.

\paragraph{Rich but efficient and effective exploration}
Unlike prior black-box hyperparameter search, the agent operates directly on the training workflow itself. Because it can read and reason about the code, it can decide which part of the pipeline to modify, whether adjusting hyperparameters, changing the learning-rate schedule, adding regularization, revising the architecture, or switching optimizers. With \texttt{python\_exec}, it can further analyze training curves, diagnose instability or overfitting, estimate feasible model sizes under compute or memory constraints, and compare evidence across runs using its persistent lessons and experimental history.

\section{Experiments}
\label{sec:experiments}
We evaluate our agent framework on 14 diverse tasks across 3 categories. In each category, we compare our general-purpose framework against the state-of-the-art methods for that task category and demonstrate its advantages, while using a comparable budget and requiring only minimal task-specific tuning beyond the domain-level system prompt template. For each setting, we conduct 3 independent runs and report the average. 
See the detailed task introduction, data split, and hyper-parameters of each setting in Appendix \ref{app:setting}.

\begin{wrapfigure}{r}{0.75\textwidth}
    \centering
    \vspace{-3mm}
    \includegraphics[
        width=\linewidth,
        trim={8mm 0 0 0}, 
        clip
    ]{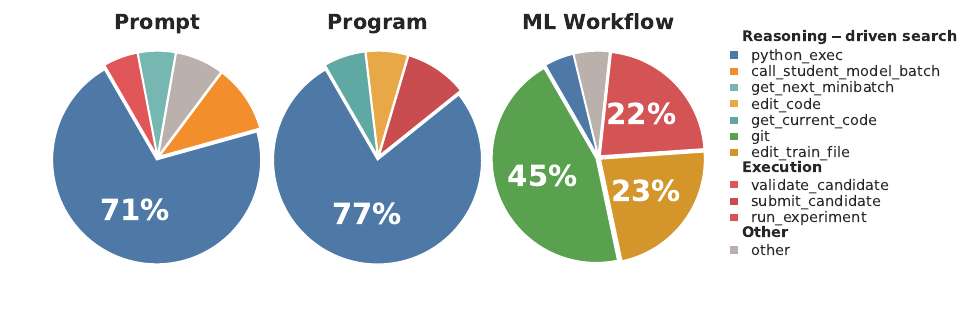}
    \vspace{-9mm}
    \caption{Tool call frequency statistics for each category.}
    \vspace{-2mm}
    \label{fig:tool pie}
\end{wrapfigure}
We first present overall tool-call frequency statistics for each category in Figure \ref{fig:tool pie}. The results highlight a characteristic pattern of ReASearch: the agent spends most of its time on reasoning rather than execution, which is typically more expensive. In prompt and program optimization, more than 70\% of tool calls are devoted to reasoning with Python. In ML workflow optimization, Python accounts for a smaller fraction of calls, but as we show later through specific examples, it remains critical to the agent's strongest results.

\subsection{Prompt optimization}

\paragraph{Settings}
The agent optimizes system prompts for student LLMs across three domains: mathematical reasoning (AIME 2025, GSM8K \citep{gsm}), multi-hop question answering (HotpotQA \citep{yang2018hotpotqa}), and software engineering (Terminal-Bench 2.0 \citep{merrill2026terminal}). The ``Baseline'' row reports the student model with the original (un-optimized) prompt. We compare our approach with the latest optimization algorithm, GEPA \citep{agrawal2025gepa}. For both methods, we use Claude Sonnet 4.6 as the agent model. As student models, we use GPT-4.1 mini for AIME and HotpotQA, Llama 3.1-8B for GSM8K, and GPT-5 for Terminal-Bench 2.0 based on the difficulties of each task. For a fair comparison, we keep the number of student-model calls the same as in GEPA. In dollar terms, a full ReASearch run costs less than \$20 in API usage with standard caching, comparable to the cost reported for GEPA in its public documentation.
The results on test sets, presented in Table \ref{tab:prompt}, show that ReASearch consistently outperforms GEPA across all tasks. 

\begin{table}[h!]
\centering
\resizebox{0.8\linewidth}{!}{%
\begin{tabular}{lcccc}
\hline
 & AIME $\uparrow$ & GSM8K $\uparrow$ & HotpotQA $\uparrow$ & Terminal-Bench 2.0 $\uparrow$ \\
\hline
Baseline   & 46.00 $\pm$ 1.33 & 81.20 $\pm$ 0.38 & 63.00 $\pm$ 1.50 & 35.56 $\pm$ 1.48 \\
GEPA       & 50.67 $\pm$ 1.15 & 82.11 $\pm$ 0.45 & 65.80 $\pm$ 0.80 & 42.22 $\pm$ 1.28 \\
ReASearch  & \textbf{52.00 $\pm$ 0.67} & \textbf{83.40 $\pm$ 0.30} & \textbf{67.60 $\pm$ 0.50} & \textbf{53.33 $\pm$ 1.96} \\
\hline
\end{tabular}%
}
\caption{Performance on test sets before and after system prompt optimization.}
\label{tab:prompt}
\end{table}

\begin{wrapfigure}{r}{0.3\linewidth}
    \centering
    \includegraphics[width=\linewidth]{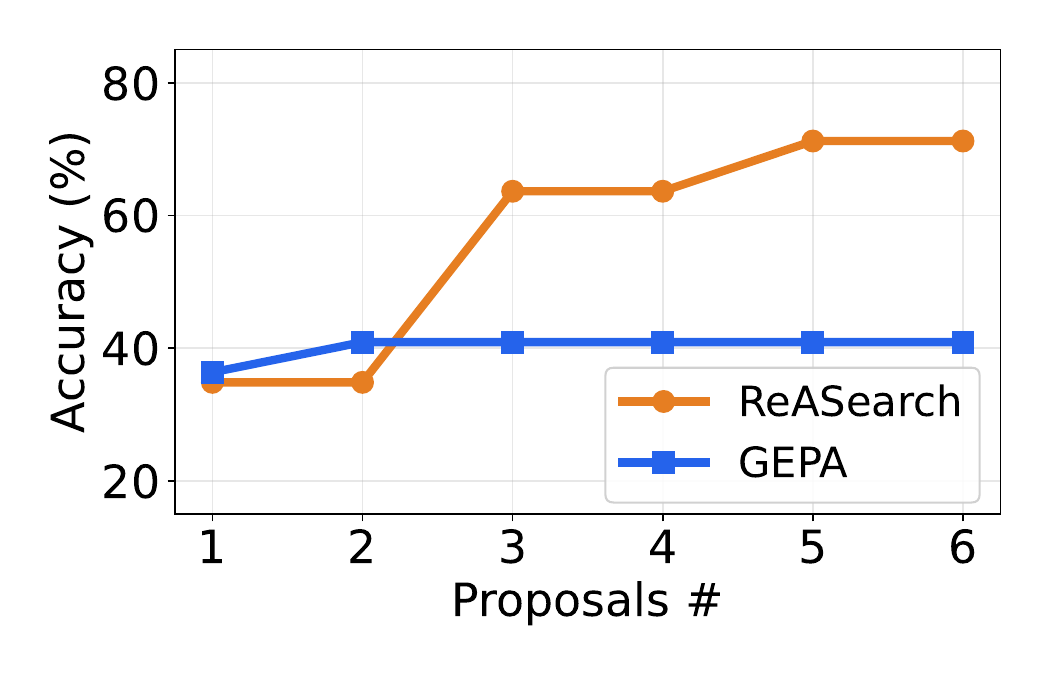}
    \vspace{-10mm}
    \caption{Comparison with GEPA on Terminal-Bench validation accuracy.}
    \label{fig:gepa}
\end{wrapfigure}
Inspecting the agent trajectories, we find that a range of optimizer-like behaviors emerge without any external controller: the agent builds a structured understanding of the task before optimizing, verifies improvements through cheap channels before committing to expensive validation, navigates a tree of candidates and reverts to earlier branches when a path fails, detects overfitting and calibrates for variance, accumulates prompt-specific knowledge in persistent memory, and selects the final prompt based on the full optimization history rather than the highest score. We detail each of these behaviors with concrete trajectory excerpts in Appendix~\ref{app:qual}.

\subsection{Program evolution}
\paragraph{Settings}
The agent optimizes Python programs for single-instance tasks across three domains: math optimization (circle packing, Heilbronn triangles), systems programming (transaction scheduling, or TXN, and expert-parallel load balancing, or EPLB), and visual reasoning (ARC-AGI-2). Each task contains challenging instances that require problem-specific test-time optimization. We compare ReASearch with prior algorithm-discovery methods under the same budget, where one budget unit is one evaluator call on the target problem; both methods are capped at the same per-task ceiling (500 for circle packing and Heilbronn per $n$, 100 per ARC-AGI-2 puzzle, 500 for EPLB and TXN). We use GPT-5 and Claude Sonnet 4.6 as the backbone models.
Tables \ref{tab:circle_packing}, \ref{tab:heilbronn}, \ref{tab:adrs}, and \ref{tab:arc_agi2} show that ReASearch delivers the best performance on nearly all tasks, in some cases even outperforming the best previously known human results, which are underlined in the tables.

\begin{table}[h]
    \centering
    \vspace{-3mm}
    \caption{Circle Packing Results}
    \label{tab:circle_packing}
    \resizebox{1.0\textwidth}{!}{
    \begin{tabular}{llcccccccccc}
    \toprule
    Algorithm & Model & $n=23$ $\uparrow$ & $n=24$ $\uparrow$ & $n=25$ $\uparrow$ & $n=26$ $\uparrow$ & $n=27$ $\uparrow$ & $n=28$ $\uparrow$ & $n=29$ $\uparrow$ & $n=30$ $\uparrow$ & $n=31$ $\uparrow$ & $n=32$ $\uparrow$ \\
    \midrule
    \multicolumn{2}{l}{Best known (human)} & 2.478 & 2.530 & 2.587 & 2.635 & 2.685 & 2.737 & 2.790 & 2.842 & 2.889 & 2.939 \\
    \midrule
    \multirow{2}{*}{AdaEvolve}
     & GPT-5      & 2.406 & 2.528 & 2.569 & 2.610 & 2.654 & 2.641 & 2.659 & 2.750 & 2.781 & 2.707 \\
     & Sonnet 4.6 & \textbf{2.478} & 2.433 & \textbf{2.587} & 2.630 & 2.576 & 2.623 & 2.786 & \underline{\textbf{2.843}} & 2.887 & 2.810 \\
    \midrule
    \multirow{2}{*}{ReASearch}
     & GPT-5      & 2.471 & 2.528 & 2.572 & 2.623 & 2.683 & 2.687 & 2.779 & 2.780 & 2.683 & 2.938 \\
     & Sonnet 4.6 & \textbf{2.478} & \textbf{2.530} & \textbf{2.587} & \underline{\textbf{2.636}} & \textbf{2.684} & \textbf{2.735} & \textbf{2.790} & \underline{\textbf{2.843}} & \underline{\textbf{2.890}} & \underline{\textbf{2.940}} \\
    \bottomrule
    \end{tabular}
    }
\end{table}

\begin{wraptable}{r}{0.64\textwidth}
    \centering
    \caption{Heilbronn Triangle Results}
    \label{tab:heilbronn}
    \resizebox{\linewidth}{!}{%
    \begin{tabular}{llccccc}
    \toprule
    Algorithm & Model & $n=11$ $\uparrow$ & $n=12$ $\uparrow$ & $n=13$ $\uparrow$ & $n=14$ $\uparrow$ & $n=15$ $\uparrow$ \\
    \midrule
    \multicolumn{2}{l}{Best known (human)} & 0.03704 & 0.03260 & 0.02702 & 0.02430 & 0.02111 \\
    \midrule
    \multirow{2}{*}{AdaEvolve}
     & GPT-5      & 0.02923 & 0.03100 & 0.02229 & 0.01606 & 0.01770 \\
     & Sonnet 4.6 & 0.02629 & 0.02552 & 0.02142 & 0.00299 & 0.00984 \\
    \midrule
    \multirow{2}{*}{ReASearch}
     & GPT-5      & 0.03539 & 0.03142 & 0.02466 & 0.02260 & 0.01903 \\
     & Sonnet 4.6 & \textbf{0.03552} & \textbf{0.03260} & \textbf{0.02700} & \textbf{0.02429} & \textbf{0.02034} \\
    \bottomrule
    \end{tabular}%
    }
    \vspace{-0.5\baselineskip}
\end{wraptable}

\begin{table}[t]
    \centering
    \vspace{-7mm}
    \begin{minipage}[t]{0.45\textwidth}
        \centering
        \caption{Systems Programming Results (TXN and EPLB)}
        \label{tab:adrs}
        \resizebox{\textwidth}{!}{%
        \begin{tabular}{llcc}
        \toprule
        Algorithm & Model & TXN $\uparrow$ & EPLB $\uparrow$ \\
        \midrule
        GEPA & GPT-5 & 3984 & 0.1445 \\
        Shinka & GPT-5 & \textbf{4329} & 0.1272 \\
        \midrule
        \multirow{2}{*}{AdaEvolve}
         & GPT-5      & 3636 & 0.1976 \\
         & Sonnet 4.6 & 4292 & 0.1441 \\
        \midrule
        \multirow{2}{*}{ReASearch}
         & GPT-5      & 4237 & \textbf{0.2305} \\
         & Sonnet 4.6 & 4032 & 0.1471 \\
        \bottomrule
        \end{tabular}%
        }
    \end{minipage}
    \hspace{0.04\textwidth}
    \begin{minipage}[t]{0.45\textwidth}
        \centering
        \caption{ARC-AGI2 Results}
        \label{tab:arc_agi2}
        \resizebox{\textwidth}{!}{%
        \begin{tabular}{llcc}
        \toprule
        Algorithm & Model & Train acc $\uparrow$ & Test acc $\uparrow$ \\
        \midrule
        \multirow{2}{*}{AdaEvolve}
         & GPT-5      & 20.8\% & 5.0\% \\
         & Sonnet 4.6 & 21.9\% & 12.5\% \\
        \midrule
        \multirow{2}{*}{ReASearch}
         & GPT-5      & 35.0\% & 11.7\% \\
         & Sonnet 4.6 & 85.0\% & \textbf{50.0\%} \\
        \bottomrule
        \end{tabular}%
        }
    \end{minipage}
\end{table}

AdaEvolve treats strategy selection as a discrete search problem: it samples paradigm labels, generates code for each, and selects the best. ReASearch gives a persistent agent full control over a multi-turn trajectory. Several capabilities explain the gap across all benchmarks:

\paragraph{Causal diagnosis drives surgical fixes.}
A persistent agent retains the full history of what was tried and why each approach stalled, so it can isolate root causes rather than regenerating code that shares the same blind spot. On EPLB, the agent had plateaued at a score of $\sim$0.21 for over 120 turns. It diagnosed that the refinement phase was stuck because the code only allowed experts with two or more replicas to donate, too restrictive to resolve many imbalances. Temporarily allowing an expert to give up its \emph{last} replica let refinement explore beneficial intermediate states it could never reach before. This single operator change broke the plateau, yielding the pivotal improvement that ultimately lifted the score from 0.21 to 0.23. AdaEvolve Sonnet, without this diagnostic capability, plateaued 61\% lower for 272 consecutive iterations; three paradigm-breakthrough waves each failed because none identified the donor constraint as the bottleneck.

\paragraph{Mathematical abstraction collapses the search space.}
Persistent context across hundreds of evaluations enables a form of scientific discovery: the agent observes patterns in its results, forms structural hypotheses, and tests them, a cycle that requires accumulating evidence over many iterations and that independent code generation cannot perform. On Heilbronn $n=12$, the agent noticed that its best point configuration had 8-fold symmetry. Rather than treating this as coincidence, it formulated the symmetry as a mathematical constraint, collapsing 24 coordinate dimensions to 2 free parameters governed by $2v^3 - v - 0.5 = 0$, and solved it to 50-digit precision. For $n=14$, after finding a candidate optimum, it verified mathematically that no local perturbation could improve the solution, establishing a provable ceiling. AdaEvolve achieved 78\% ($n=12$) and 12\% ($n=14$): without multi-turn agentic reasoning and persistent memory, baseline methods fail to abstract scattered observations into formal mathematical assumptions. On transaction scheduling, the same capacity led the agent to diagnose that its 1752 lines of metaheuristics spent the 16-second budget on search machinery rather than on the fast moves that actually reduce cost, and to strip the code down to 244 lines, achieving a +51\% improvement.

\paragraph{Composite algorithms emerge from iterative discovery.}
Effective optimization often requires combining multiple techniques, each addressing a different subproblem. Discovering such composites means testing components in isolation before designing their integration, a sequential dependency that single-pass code generation, which must invent everything at once, cannot easily satisfy. The circle packing agent found that computing optimal radii for fixed circle centers reduces to a linear program solvable in $\sim$4ms. After verifying this independently, it embedded the LP as an inner loop inside stochastic position search, so each candidate layout receives guaranteed-optimal radii at negligible cost. This hybrid outperformed both components alone and emerged naturally from a multi-turn process where each design step built on the verified outcome of the previous one.

\paragraph{Analyze first, code second.}
ReASearch shows the most improvement on ARC-AGI-2, where the task is to infer transformation rules from few examples. Besides stronger reasoning, we attribute the 4$\times$ test accuracy gap to the agent's tool-mediated analysis cycle: using \texttt{python\_exec} to interrogate training pairs before writing code, and using evaluator feedback to diagnose errors cell by cell after submission. On Task~72 (symmetry with minimum removals), 31 turns of analysis, decomposing shapes, testing LR vs.\ UD vs.\ rot180 symmetry, computing removal sets, comparing with expected outputs, confirmed the rule is specifically LR symmetry with integer-center tie-breaking. AdaEvolve's best program on the same task searches all three symmetry types and picks whichever globally retains the most cells; because it never verified which type the task requires, it sometimes selects a UD axis, achieving 99.5\% cell accuracy but only 2 of 4 training examples exactly correct. On Task~44 (beam tracing), 13 turns of analysis identified that a pointer shoots a beam through shapes that redirect it via 3-markers. AdaEvolve's best program encodes a fundamentally different model (static axis with perpendicular fills) that achieves 96\% cell accuracy but is structurally wrong. Across 22 tasks where ReASearch scored perfectly on test but AdaEvolve fell below 0.5, the pattern was the same: AdaEvolve's code implemented an approximately-right model (90--99\% cell accuracy) whose remaining errors reflected a wrong structural assumption that became fatal on unseen inputs.

\subsection{ML workflow optimization} 
\paragraph{Settings}
We compare our ReASearch scaffold against Claude Code across several ML domains, including language modeling (NanoGPT), 100-class image classification (IMG-100), reinforcement learning (Atari Q*bert and MuJoCo), and time series forecasting (the DRW Crypto Market Prediction Kaggle competition).
We use bits per byte (BPB), accuracy, and reward as the evaluation metrics for the first four tasks, and the correlation coefficient for the crypto prediction task. For the latter, we report the score of the submitted test CSV on the private split to mirror the real Kaggle competition setting.
Each experiment is run under a fixed per-experiment training time budget (5--30 minutes depending on the task), and both ReASearch and Claude Code are given the same total wall-clock budget per task. The Claude Code baseline is the official AutoResearch setup with Claude Sonnet 4.6, used unmodified except to match our wall-clock budget; the training script, data pipeline, evaluator, hardware, and per-experiment training cap are identical on both sides. See Appendix~\ref{app:claude_code} for the matched-instruction comparison and an analysis of where the score gap comes from.

\begin{figure}
    \centering
    \vspace{-12mm}
    \begin{minipage}[t]{0.38\linewidth}
        \centering
        \includegraphics[width=\linewidth]{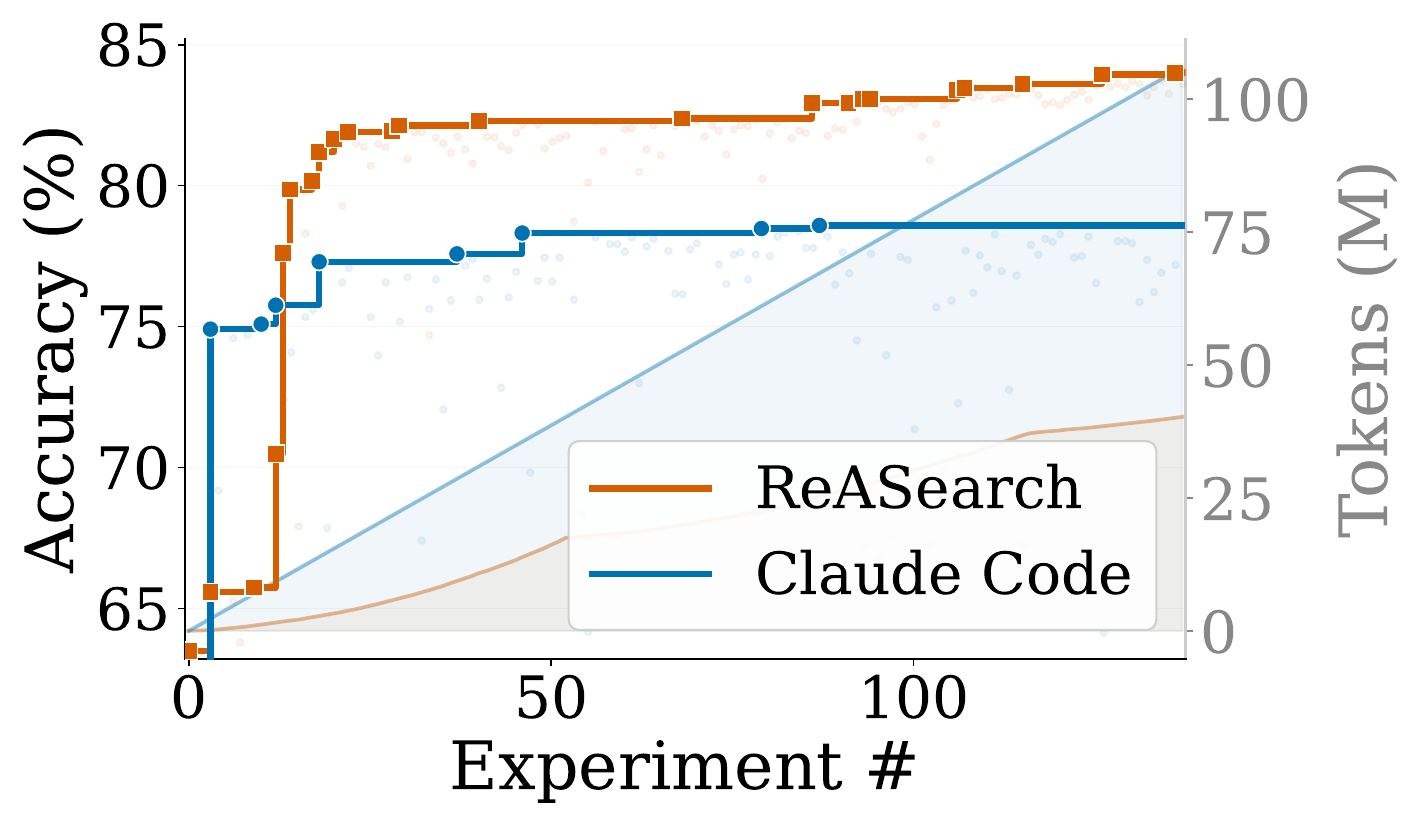}
    \end{minipage}\hfill
    \begin{minipage}[t]{0.38\linewidth}
        \centering
        \includegraphics[width=\linewidth]{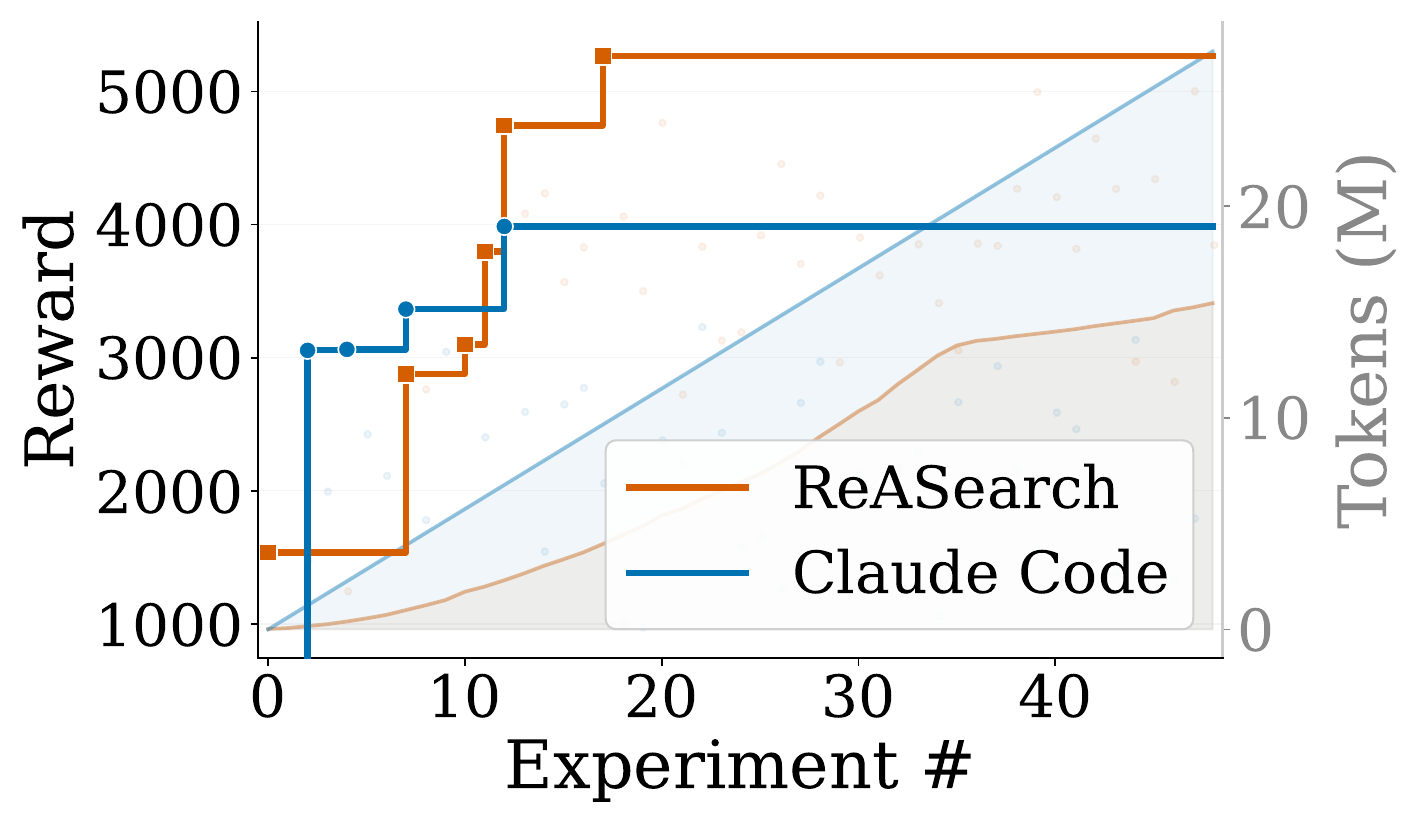}
    \end{minipage}
    \vspace{-3mm}
    \caption{Performance and token usage across experiments. For Claude Code, token usage is only available at the end of the optimization run, so we plot it as a linear trajectory.}
    \label{fig:ml}
\end{figure}

We report the best result achieved under the same experiment budget in Table \ref{tab:ml}, together with the corresponding optimization curves in Figure \ref{fig:ml}. ReASearch attains stronger performance on most training tasks while using far fewer tokens. The two systems are statistically indistinguishable on NanoGPT (0.976 $\pm$ 0.008 vs.\ 0.974 $\pm$ 0.010), but ReASearch shows a clear advantage on the remaining tasks. By contrast, Claude Code sometimes becomes trapped in unproductive regions of the search space and explores less efficiently. As detailed in Appendix~\ref{app:claude_code}, which describes the matched-instruction comparison design, the gap stems from a harness that re-emits the current search state (best result, recent experiments, persistent lessons, and stagnation warnings) into the prompt every turn, rather than from a difference in the instructions given to the two systems. Notably, within only 15 experiments, ReASearch finds a configuration that improves the submission's Kaggle leaderboard ranking from 36th to 6th (starting from a baseline that already incorporates a small amount of manual iteration; see Appendix~\ref{app:setting}). This result highlights the practical potential of our system in realistic settings, where test labels are hidden and progress must be made under limited evaluation opportunities.

\begin{table}[h!]
\centering
\resizebox{\linewidth}{!}{%
\begin{tabular}{lccccc}
\hline
Method & NanoGPT $\downarrow$ & IMG-100 $\uparrow$ & Atari (Q*bert) $\uparrow$ & MuJoCo $\uparrow$ & Crypto $\uparrow$ \\
\hline
Baseline    & 0.998 $\pm$ 0.006 & 63.51 $\pm$ 0.85 & 475 $\pm$ 90 & 1537 $\pm$ 220 & 0.0953 $\pm$ 0.0021 (rank 36) \\
Claude Code & \textbf{0.974 $\pm$ 0.010} & 78.59 $\pm$ 1.40 & 1250 $\pm$ 180 & 3986 $\pm$ 410 & 0.0999 $\pm$ 0.0028 (rank 29) \\
ReASearch   & 0.976 $\pm$ 0.008 & \textbf{83.99 $\pm$ 1.10} & \textbf{4500 $\pm$ 320} & \textbf{5267 $\pm$ 480} & \textbf{0.1110 $\pm$ 0.0024 (rank 6)} \\
\hline
\end{tabular}%
}
\caption{Model training tasks.}
\vspace{-4mm}
\label{tab:ml}
\end{table}

\paragraph{Phased search with constraint-aware reasoning.}
Across all domains the agent consistently exhibits a multi-phase optimization strategy that mirrors how human researchers approach experimental campaigns, following a three-phase pattern: (i)~broad architectural exploration, (ii)~focused hyperparameter refinement, and (iii)~diminishing-returns fine-tuning. In IMG-100 under a 5-minute training budget, the agent first swept architectures, including WideResNet-28-10 (53.9\%, too memory-heavy) and RandAugment (60.97\%, too slow to converge), settling on ResNet-18 with CutMix at 65.75\%. It then used \texttt{python\_exec} to estimate wall-clock time per step (${\sim}$0.03--0.05\,s), calculated that only ${\sim}$60--100 epochs would complete in 300 seconds, and noticed the cosine learning-rate schedule was set to decay over 200 epochs, so training always stopped while the learning rate was still high. Sweeping \texttt{estimated\_epochs} confirmed the diagnosis: setting it to 100 (matching the actual epoch count) jumped accuracy from 65.75\% to 79.86\%, a +14 percentage-point gain from a single insight. Claude Code never performed this calculation and stumbled onto OneCycleLR only after dozens of wasted experiments. In Atari Q*bert, a similar pattern emerged: after scaling from 8 to 16 parallel environments, the agent
compared throughput (78$\to$247 steps/sec) against reward (350 both times) and concluded that \emph{learning efficiency, not data quantity} was the bottleneck, leading it to deepen the FC layers (512$\to$1024$\to$512) for a 5$\times$ reward jump, while Claude Code ran 90 experiments on CNN architectures without ever testing this.

\paragraph{Diagnostic learning from failure.}
Rather than merely blacklisting failed configurations, the agent uses \texttt{python\_exec} to extract \emph{causal} explanations that redirect subsequent search. In MuJoCo, when prioritized experience replay (PER) was proposed as a standard RL improvement, the agent first benchmarked \texttt{np.random.choice} with priority weights against uniform \texttt{np.random.randint}, measured a 133$\times$ slowdown, and rejected PER on computational grounds, saving an experiment that would have timed out (as Claude Code later confirmed). In IMG-100, two early EMA (exponential moving average) attempts collapsed to ${\sim}$1\% accuracy. Rather than abandoning EMA, the agent traced the failure to batch-normalization running statistics not being propagated through naive weight averaging, then fixed it using PyTorch's \texttt{AveragedModel} with post-training BN recomputation, recovering a +0.47\% gain (83.47\%$\to$83.94\%). Claude Code encountered the same EMA failure and never revisited it.

\paragraph{Discovery of co-dependent improvements.}
The agent's most distinctive capability is detecting when parameter changes are \emph{co-dependent}, i.e., improvements that only materialize when applied together, which single-parameter sweeps systematically miss. In IMG-100, AMP (mixed-precision training) appeared to be a marginal speed optimization in isolation, but the agent recognized it as a \emph{VRAM enabler}: reducing memory from 14.3\,GB to 4.1\,GB made deeper architectures feasible, which in turn made CutMix augmentation beneficial (enough epochs to help rather than hurt) and EMA stable (enough capacity to regularize). This chain of co-dependent gains (82.38\%$\to$84.03\%) collapsed if any link was removed, as Claude Code demonstrated by never discovering AMP's enabling role, capping at 78.59\%. In NanoGPT, the agent used \texttt{python\_exec} to sort all experiments by validation loss and identify ``near-miss'' improvements small enough to be independent, then combined them; critically, it discovered that reducing batch size from $2^{18}$ to $2^{17}$ required simultaneously re-tuning the learning rate (0.04$\to$0.027) and warmdown schedule (0.6715$\to$0.702), as applying the old hyperparameters to the new batch size would have been \emph{worse} than the original configuration, a trap Claude Code partially fell into by not propagating changes through dependent parameters.

\subsection{Ablation studies}
We ablate ReASearch's core components (memory and the Python execution tool), memory design and transfer, and validation-feedback granularity. Both components matter, with memory dominating on long-trajectory prompt tasks and Python execution on ARC-AGI-2; a simple \emph{what worked / what did not / what to try next} schema suffices and transfers across runs; and exposing only aggregate validation metrics avoids overfitting. Full results are in Appendix~\ref{app:ablation}.

\section{Conclusion}
We propose ReASearch, a unified framework that turns optimization itself into a reasoning problem. By replacing hand-designed search logic with agentic reasoning over tools, ReASearch enables a single system to optimize prompts, programs, and ML workflows with strong empirical gains. These results suggest a broader view: reasoning is not just for solving tasks, but can itself be a powerful engine for open-ended search.

\bibliography{ref}

@inproceedings{opro,
  title={Large language models as optimizers},
  author={Yang, Chengrun and Wang, Xuezhi and Lu, Yifeng and Liu, Hanxiao and Le, Quoc V and Zhou, Denny and Chen, Xinyun},
  booktitle={The Twelfth International Conference on Learning Representations},
  year={2023}
}

@article{khattab2023dspy,
  title={Dspy: Compiling declarative language model calls into self-improving pipelines},
  author={Khattab, Omar and Singhvi, Arnav and Maheshwari, Paridhi and Zhang, Zhiyuan and Santhanam, Keshav and Vardhamanan, Sri and Haq, Saiful and Sharma, Ashutosh and Joshi, Thomas T and Moazam, Hanna and others},
  journal={arXiv preprint arXiv:2310.03714},
  year={2023}
}

@article{yuksekgonul2024textgrad,
  title={Textgrad: Automatic" differentiation" via text},
  author={Yuksekgonul, Mert and Bianchi, Federico and Boen, Joseph and Liu, Sheng and Huang, Zhi and Guestrin, Carlos and Zou, James},
  journal={arXiv preprint arXiv:2406.07496},
  year={2024}
}

@article{adalflow,
  title={Llm-autodiff: Auto-differentiate any llm workflow},
  author={Yin, Li and Wang, Zhangyang},
  journal={arXiv preprint arXiv:2501.16673},
  year={2025}
}

@article{novikov2025alphaevolve,
  title={Alphaevolve: A coding agent for scientific and algorithmic discovery},
  author={Novikov, Alexander and V{\~u}, Ng{\^a}n and Eisenberger, Marvin and Dupont, Emilien and Huang, Po-Sen and Wagner, Adam Zsolt and Shirobokov, Sergey and Kozlovskii, Borislav and Ruiz, Francisco JR and Mehrabian, Abbas and others},
  journal={arXiv preprint arXiv:2506.13131},
  year={2025}
}

@article{cemri2026adaevolve,
  title={AdaEvolve: Adaptive LLM driven zeroth-order optimization},
  author={Cemri, Mert and Agrawal, Shubham and Gupta, Akshat and Liu, Shu and Cheng, Audrey and Mang, Qiuyang and Naren, Ashwin and Erdogan, Lutfi Eren and Sen, Koushik and Zaharia, Matei and others},
  journal={arXiv preprint arXiv:2602.20133},
  year={2026}
}

@article{ding2025scaling,
  title={Scaling Textual Gradients via Sampling-Based Momentum},
  author={Ding, Zixin and Hong, Junyuan and Shi, Zhan and Wang, Jiachen T and Lin, Zinan and Yin, Li and Liu, Meng and Wang, Zhangyang and Chen, Yuxin},
  journal={arXiv preprint arXiv:2506.00400},
  year={2025}
}

@article{wang2023promptagent,
  title={Promptagent: Strategic planning with language models enables expert-level prompt optimization},
  author={Wang, Xinyuan and Li, Chenxi and Wang, Zhen and Bai, Fan and Luo, Haotian and Zhang, Jiayou and Jojic, Nebojsa and Xing, Eric P and Hu, Zhiting},
  journal={arXiv preprint arXiv:2310.16427},
  year={2023}
}

@inproceedings{mipro,
  title={Optimizing instructions and demonstrations for multi-stage language model programs},
  author={Opsahl-Ong, Krista and Ryan, Michael J and Purtell, Josh and Broman, David and Potts, Christopher and Zaharia, Matei and Khattab, Omar},
  booktitle={Proceedings of the 2024 Conference on Empirical Methods in Natural Language Processing},
  pages={9340--9366},
  year={2024}
}

@article{agrawal2025gepa,
  title={GEPA: Reflective prompt evolution can outperform reinforcement learning},
  author={Agrawal, Lakshya A and Tan, Shangyin and Soylu, Dilara and Ziems, Noah and Khare, Rishi and Opsahl-Ong, Krista and Singhvi, Arnav and Shandilya, Herumb and Ryan, Michael J and Jiang, Meng and others},
  journal={arXiv preprint arXiv:2507.19457},
  year={2025}
}

@inproceedings{ape,
  title={Large language models are human-level prompt engineers},
  author={Zhou, Yongchao and Muresanu, Andrei Ioan and Han, Ziwen and Paster, Keiran and Pitis, Silviu and Chan, Harris and Ba, Jimmy},
  booktitle={The eleventh international conference on learning representations},
  year={2022}
}

@software{openevolve,
  title = {OpenEvolve: an open-source evolutionary coding agent},
  author = {Asankhaya Sharma},
  year = {2025},
  publisher = {GitHub},
  url = {https://github.com/algorithmicsuperintelligence/openevolve}
}

@article{lange2025shinkaevolve,
  title={Shinkaevolve: Towards open-ended and sample-efficient program evolution},
  author={Lange, Robert Tjarko and Imajuku, Yuki and Cetin, Edoardo},
  journal={arXiv preprint arXiv:2509.19349},
  year={2025}
}

@article{assumpccao2025codeevolve,
  title={Codeevolve: An open source evolutionary coding agent for algorithm discovery and optimization},
  author={Assump{\c{c}}{\~a}o, Henrique and Ferreira, Diego and Campos, Leandro and Murai, Fabricio},
  journal={arXiv preprint arXiv:2510.14150},
  year={2025}
}

@inproceedings{yang2018hotpotqa,
  title={HotpotQA: A dataset for diverse, explainable multi-hop question answering},
  author={Yang, Zhilin and Qi, Peng and Zhang, Saizheng and Bengio, Yoshua and Cohen, William and Salakhutdinov, Ruslan and Manning, Christopher D},
  booktitle={Proceedings of the 2018 conference on empirical methods in natural language processing},
  pages={2369--2380},
  year={2018}
}

@article{gsm,
  title={Training verifiers to solve math word problems},
  author={Cobbe, Karl and Kosaraju, Vineet and Bavarian, Mohammad and Chen, Mark and Jun, Heewoo and Kaiser, Lukasz and Plappert, Matthias and Tworek, Jerry and Hilton, Jacob and Nakano, Reiichiro and others},
  journal={arXiv preprint arXiv:2110.14168},
  year={2021}
}

@article{merrill2026terminal,
  title={Terminal-bench: Benchmarking agents on hard, realistic tasks in command line interfaces},
  author={Merrill, Mike A and Shaw, Alexander G and Carlini, Nicholas and Li, Boxuan and Raj, Harsh and Bercovich, Ivan and Shi, Lin and Shin, Jeong Yeon and Walshe, Thomas and Buchanan, E Kelly and others},
  journal={arXiv preprint arXiv:2601.11868},
  year={2026}
}

@article{heilbronn,
  title={From Computational Certification to Exact Coordinates: Heilbronn's Triangle Problem on the Unit Square Using Mixed-Integer Optimization},
  author={Sudermann-Merx, Nathan},
  journal={arXiv preprint arXiv:2603.11107},
  year={2026}
}

@article{chollet2025arc,
  title={Arc-agi-2: A new challenge for frontier ai reasoning systems},
  author={Chollet, Francois and Knoop, Mike and Kamradt, Gregory and Landers, Bryan and Pinkard, Henry},
  journal={arXiv preprint arXiv:2505.11831},
  year={2025}
}

@misc{cursor2025,
  author = {{Anysphere, Inc.}},
  title = {Cursor},
  year = {2025},
  url = {https://cursor.com/home},
}

@misc{claudecode2025,
  author = {{Anthropic}},
  title = {Claude Code},
  year = {2025},
  url = {https://claude.com/product/claude-code},
}

@misc{opencode2025,
  author = {{OpenCode}},
  title = {OpenCode},
  year = {2025},
  url = {https://opencode.ai/},
}

@misc{codex2025,
  author = {{OpenAI}},
  title = {Codex},
  year = {2025},
  url = {https://openai.com/codex/},
}

@article{kang2025textbo,
  title={TextBO: Bayesian Optimization in Language Space for Eval-Efficient Self-Improving AI},
  author={Kang, Enoch H and Yoganarasimhan, Hema},
  year={2025}
}

@article{schneider2024hyperband,
  title={Hyperband-based Bayesian optimization for black-box prompt selection},
  author={Schneider, Lennart and Wistuba, Martin and Klein, Aaron and Golebiowski, Jacek and Zappella, Giovanni and Merra, Felice Antonio},
  journal={arXiv preprint arXiv:2412.07820},
  year={2024}
}

@article{bergstra2011algorithms,
  title={Algorithms for hyper-parameter optimization},
  author={Bergstra, James and Bardenet, R{\'e}mi and Bengio, Yoshua and K{\'e}gl, Bal{\'a}zs},
  journal={Advances in neural information processing systems},
  volume={24},
  year={2011}
}

@article{shi2024efficient,
  title={Efficient prompt optimization through the lens of best arm identification},
  author={Shi, Chengshuai and Yang, Kun and Chen, Zihan and Li, Jundong and Yang, Jing and Shen, Cong},
  journal={Advances in Neural Information Processing Systems},
  volume={37},
  pages={99646--99685},
  year={2024}
}

@inproceedings{pryzant2023automatic,
  title={Automatic prompt optimization with “gradient descent” and beam search},
  author={Pryzant, Reid and Iter, Dan and Li, Jerry and Lee, Yin and Zhu, Chenguang and Zeng, Michael},
  booktitle={Proceedings of the 2023 conference on empirical methods in natural language processing},
  pages={7957--7968},
  year={2023}
}

@article{li2026efficient,
  title={Efficient Multi-objective Prompt Optimization via Pure-exploration Bandits},
  author={Li, Donghao and Shi, Chengshuai and Ou, Weijuan and Shen, Cong and Yang, Jing},
  journal={arXiv preprint arXiv:2605.14553},
  year={2026}
}

@article{wu2025llm,
  title={Llm prompt duel optimizer: Efficient label-free prompt optimization},
  author={Wu, Yuanchen and Verma, Saurabh and Lee, Justin and Xiong, Fangzhou and Zhang, Poppy and Awadelkarim, Amel and Chen, Xu and Yuan, Yubai and Hill, Shawndra},
  journal={arXiv preprint arXiv:2510.13907},
  year={2025}
}

@article{guo2023evoprompt,
  title={Evoprompt: Connecting llms with evolutionary algorithms yields powerful prompt optimizers},
  author={Guo, Qingyan and Wang, Rui and Guo, Junliang and Li, Bei and Song, Kaitao and Tan, Xu and Liu, Guoqing and Bian, Jiang and Yang, Yujiu},
  journal={arXiv e-prints},
  pages={arXiv--2309},
  year={2023}
}

@article{fernando2023promptbreeder,
  title={Promptbreeder: Self-referential self-improvement via prompt evolution},
  author={Fernando, Chrisantha and Banarse, Dylan and Michalewski, Henryk and Osindero, Simon and Rockt{\"a}schel, Tim},
  journal={arXiv preprint arXiv:2309.16797},
  year={2023}
}

@article{secheresse2025gaapo,
  title={GAAPO: genetic algorithmic applied to prompt optimization},
  author={S{\'e}cheresse, Xavier and Guilbert--Ly, Jacques-Yves and Villedieu de Torcy, Antoine},
  journal={Frontiers in Artificial Intelligence},
  volume={8},
  pages={1613007},
  year={2025},
  publisher={Frontiers Media SA}
}

@article{zelikman2023self,
  title={Self-taught optimizer (stop): Recursively self-improving code generation},
  author={Zelikman, Eric and Lorch, Eliana and Mackey, Lester and Kalai, Adam Tauman},
  journal={arXiv preprint arXiv:2310.02304},
  year={2023}
}

@inproceedings{shi2026generalizable,
  title={Generalizable Heuristic Generation Through LLMs with Meta-Optimization},
  author={Shi, Yiding and Zhou, Jianan and Song, Wen and Bi, Jieyi and Wu, Yaoxin and Cao, Zhiguang and Zhang, Jie},
  booktitle={The Fourteenth International Conference on Learning Representations},
  year={2026}
}

@article{van2024llamea,
  title={Llamea: A large language model evolutionary algorithm for automatically generating metaheuristics},
  author={Van Stein, Niki and B{\"a}ck, Thomas},
  journal={IEEE Transactions on Evolutionary Computation},
  volume={29},
  number={2},
  pages={331--345},
  year={2024},
  publisher={IEEE}
}

@inproceedings{hu2025automated,
  title={Automated design of agentic systems},
  author={Hu, Shengran and Lu, Cong and Clune, Jeff},
  booktitle={International Conference on Learning Representations},
  volume={2025},
  pages={21344--21377},
  year={2025}
}

@article{lin2026agentic,
  title={Agentic harness engineering: Observability-driven automatic evolution of coding-agent harnesses},
  author={Lin, Jiahang and Liu, Shichun and Pan, Chengjun and Lin, Lizhi and Dou, Shihan and Huang, Xuanjing and Yan, Hang and Han, Zhenhua and Gui, Tao},
  journal={arXiv preprint arXiv:2604.25850},
  year={2026}
}

@article{zhang2025darwin,
  title={Darwin godel machine: Open-ended evolution of self-improving agents},
  author={Zhang, Jenny and Hu, Shengran and Lu, Cong and Lange, Robert and Clune, Jeff},
  journal={arXiv preprint arXiv:2505.22954},
  year={2025}
}

@article{jiang2026deltaevolve,
  title={DeltaEvolve: Accelerating Scientific Discovery through Momentum-Driven Evolution},
  author={Jiang, Jiachen and Ding, Tianyu and Zhu, Zhihui},
  journal={arXiv preprint arXiv:2602.02919},
  year={2026}
}

@article{yan2026pacevolve,
  title={Pacevolve: Enabling long-horizon progress-aware consistent evolution},
  author={Yan, Minghao and Peng, Bo and Coleman, Benjamin and Chen, Ziqi and Xie, Zhouhang and Chen, Shuo and He, Zhankui and Sachdeva, Noveen and Ye, Isabella and Wang, Weili and others},
  journal={arXiv preprint arXiv:2601.10657},
  year={2026}
}

@article{ye2026evaluation,
  title={Evaluation-driven scaling for scientific discovery},
  author={Ye, Haotian and Lin, Haowei and Tang, Jingyi and Luo, Yizhen and Yang, Caiyin and Su, Chang and Thapa, Rahul and Yang, Rui and Liu, Ruihua and Li, Zeyu and others},
  journal={arXiv preprint arXiv:2604.19341},
  year={2026}
}

@article{liu2602evox,
  title={Evox: Meta-evolution for automated discovery},
  author={Liu, Shu and Agarwal, Shubham and Maheswaran, Monishwaran and Cemri, Mert and Li, Zhifei and Mang, Qiuyang and Naren, Ashwin and Boneh, Ethan and Cheng, Audrey and Pan, Melissa Z and others},
  journal={arXiv preprint arXiv:2602.23413},
  year={2026}
}

@misc{autoresearch,
  author = {Andrej Karpathy},
  title = {AutoResearch},
  year = {2025},
  publisher = {GitHub},
  url = {https://github.com/karpathy/autoresearch}
}

@misc{autoresearchathome,
  author = {{Mutable State, Inc.}},
  title = {AutoResearch-at-home},
  year = {2025},
  publisher = {GitHub},
  url = {https://github.com/mutable-state-inc/autoresearch-at-home}
}

@misc{friedmanpacking,
  author = {Erich Friedman},
  title = {Circles in Squares (packing best-known results)},
  year = {2009},
  howpublished = {\url{https://erich-friedman.github.io/packing/cirRsqu/}}
}

@misc{deepseekeplb,
  author = {{DeepSeek-AI}},
  title = {EPLB: Expert Parallelism Load Balancer},
  year = {2025},
  publisher = {GitHub},
  url = {https://github.com/deepseek-ai/eplb}
}

@misc{nanogpt,
  author = {Andrej Karpathy},
  title = {nanoGPT},
  year = {2022},
  publisher = {GitHub},
  url = {https://github.com/karpathy/nanogpt}
}
\bibliographystyle{colm2026_conference}

\newpage
\appendix

\section{Related work}
\label{app:related}

\paragraph{LLM-based text optimization.}
Prior work on LLM-based text optimization typically relies on curated external heuristics to govern the overall optimization pipeline, including APE \citep{ape}, OPRO \citep{opro}, ADAS \citep{hu2025automated}, DGM \citep{zhang2025darwin}, AlphaEvolve \citep{novikov2025alphaevolve, openevolve}, CodeEvolve \citep{assumpccao2025codeevolve}, PACEvolve \citep{yan2026pacevolve}, DeltaEvolve \citep{jiang2026deltaevolve}, AHE \citep{lin2026agentic}, and SimpleTES \citep{ye2026evaluation}.
These methods introduce a range of algorithmic heuristics to guide search, such as textual-gradient-based backpropagation in DSPy, TextGrad, and AdalFlow \citep{khattab2023dspy, yuksekgonul2024textgrad, adalflow}; momentum-based updates \citep{ding2025scaling}; MCTS-based exploration in PromptAgent  \citep{wang2023promptagent}; Bayesian optimization as in MIPRO \citep{mipro,kang2025textbo}; bandit-based allocation in ShinkaEvolve and AdaEvolve \citep{lange2025shinkaevolve,cemri2026adaevolve}; Pareto-frontier-based selection in GEPA \citep{agrawal2025gepa}; and adaptive mathematical heuristics in STOP \citep{zelikman2023self}, LLaMEA \citep{van2024llamea}, MoH \citep{shi2026generalizable} and EvoX \citep{liu2602evox}.
Despite their algorithmic diversity, a unifying limitation of these methods is their reliance on an external controller. The LLM functions merely as a stateless mutation or reflection subroutine, while the actual search policy, such as budget allocation, population management, and exploration-exploitation tradeoff, is hard-coded externally.

\paragraph{Autonomous ML Agents.}
More recently, autonomous agent frameworks such as AutoResearch \citep{autoresearch, autoresearchathome} are closer in spirit to our approach. However, these systems primarily serve as single-domain proof-of-concepts tailored specifically for ML training pipelines, and often rely on closed-source orchestrations like Claude Code that obscure the underlying search dynamics. Crucially, they leave a fundamental empirical question unanswered: to what extent can agentic reasoning serve as a universal optimization policy? Our work eliminates the external controller entirely and systematically demonstrates that a single, minimalist agent scaffold can internalize complex search behaviors across three fundamentally distinct optimization modalities (prompts, programs, and ML workflows).

\paragraph{Code agents.}
A parallel line of work equips LLMs with tools such as file editing, code execution, and terminal access for software engineering tasks, as seen in systems such as Cursor \citep{cursor2025}, Claude Code \citep{claudecode2025}, OpenCode \citep{opencode2025}, and Codex \citep{codex2025}. While these agents excel at multi-step reasoning over codebases for open-ended development, they are not inherently architected for rigid, budget-constrained optimization loops. In our empirical evaluations, we directly benchmark ReASearch against a Claude Code baseline. We demonstrate that while generalist code agents can execute experimental pipelines, they frequently struggle with constraint-aware reasoning and fall into local optima compared to a reasoning-driven scaffold purpose-built for optimization. By retrofitting tool-augmented code agents into formal optimization settings, our work proves that agentic reasoning can successfully replace hand-designed mathematical meta-heuristics.

\section{Extended qualitative analysis of prompt optimization}
\label{app:qual}

We expand on the optimizer-like behaviors summarized in Section~\ref{sec:experiments}, providing concrete trajectory excerpts for each.

\paragraph{The agent builds a structured understanding of the task before optimizing.}
Rather than immediately generating prompt variants, the optimizer agents first invest in understanding \emph{what kinds of problems the student faces and why}. On HotpotQA, the agent used \texttt{python\_exec} to parse the baseline evaluation log and categorized all 126 failures into a taxonomy (categories overlap, as a single failure can exhibit more than one mode): 57 verbosity cases (the student answered ``No, only Terence Fisher was a British film director'' instead of simply ``no''), 68 wrong-reasoning cases, and 8 yes/no answers polluted with explanations. This taxonomy directly shaped the prompt: the dominant verbosity mode led to the ``minimum phrase'' framing that drove accuracy to 66.0\%. Crucially, when designing subsequent candidates, the agent \emph{recalled} specific previously-failing examples to verify that new fixes addressed them, e.g., checking whether ID~260 (where the student confused a company with its CEO) now returned a person's name after adding the rule ``if asked `who,' answer with a person, not an organization.''

\paragraph{The agent verifies improvements before committing to expensive validation.}
Validation is the costliest operation in the budget, so the agents build confidence through cheaper channels first: inspecting student outputs on train batches, comparing error patterns across candidates, and confirming improvements happen for the \emph{right reasons}. On AIME, after designing Candidate~\#5 with targeted math heuristics, the agent used \texttt{python\_exec} to compare all 15 remaining errors against the previous candidate's 17, verifying that the anti-hallucination rule had eliminated ``known answer'' citations and that casework guidance reduced counting errors. Only after this content-level analysis confirmed the prompt was improving for the right reasons did the agent commit to validation.

\paragraph{The agent autonomously navigates a tree of candidates, reverting to earlier branches when a path fails.}
Without any external controller prescribing when to backtrack, the agents maintain awareness of candidate lineage and independently decide to restart from a known-good ancestor. On Terminal-Bench, after the strategy-first prompt (Candidate~\#3, 63.6\%) was extended with three bundled refinements in Candidate~\#4, validation dropped by 3 points despite encouraging training scores. The agent reasoned about \emph{what specifically caused the regression} (a narrower search command missed files the original had caught), reverted to Candidate~\#3's exact text, and added back only one sentence about heredoc formatting. This targeted addition (Candidate~\#5) reached 71.2\%, the run's best. The revert was deliberate: the agent returned to a specific node in the search tree because it understood which branch introduced the problem.

\paragraph{The agent detects overfitting and calibrates for variance.}
On AIME, the agent enriched the baseline with domain-specific math heuristics and watched training accuracy climb from 60\% to 66.7\%, but validation returned 50.2\%, \emph{below} the 51.1\% baseline. Rather than trying more heuristics, the agent recorded a permanent lesson (``domain-specific additions hurt validation generalization even when they help train'') and reversed course, spending the rest of the run \emph{removing} complexity. On HotpotQA, the agent re-ran the best candidate's validation and obtained a score 4.2 points lower, immediately recalibrating: ``we need consistent improvements of 3--5 points to be confident any change is real.''

\paragraph{The agent accumulates prompt-specific knowledge in persistent memory.}
In the writable \texttt{lessons.md} file, the agents automatically record which prompt elements help or hurt which examples, which modifications have been tried and failed, and what structural properties they have discovered about how the student model processes instructions. This accumulated knowledge directly guides later decisions: on AIME, the recorded lesson ``domain-specific additions hurt validation generalization even when they help train'' reversed the agent's entire strategy from enriching prompts to simplifying them; on HotpotQA, its record of conflicting rules (``use full names'' vs.\ ``don't repeat question nouns'') prevented it from revisiting a dead-end direction after context compression.

\paragraph{The agent selects the final prompt based on the full optimization history, not just the highest score.}
On AIME, the agent made a bolder bet: although a simple baseline remained one of the best-\emph{validated} prompts after 16 candidates, the agent submitted a new synthesis combining the best-training candidate's strengths with targeted fixes, stripped of the additions it had identified as causes of overfitting. This unvalidated prompt scored 52.0\% on the 150-example test set, outperforming the best-validated prompt's 49.3\%, suggesting that accumulated understanding of what generalizes provided a more reliable signal than validation scores alone.

\section{Ablation studies}
\label{app:ablation}

\paragraph{Component attribution.}
We isolate the contribution of two core components of ReASearch by removing them while keeping all other settings unchanged: (1) the memory mechanism (lesson summaries and the search tree) and (2) the Python execution tool. Tables~\ref{tab:ablation_prompt} and~\ref{tab:ablation_arc} report results on representative prompt-optimization and program-evolution tasks. Both components are important, but their relative impact varies across tasks. For prompt optimization on AIME and Terminal-Bench, the agent reads and analyzes long evaluation trajectories to identify weaknesses in the current prompt; these findings are easily lost across iterations, so memory has the larger impact. For ARC-AGI-2, where interaction rounds are shorter but pattern recognition and algorithmic discovery are central, Python execution has the larger impact, since it lets the agent reason about candidate algorithms and test them before committing. Heilbronn ablation results follow the same pattern and are reported in Appendix~\ref{app:ablation_heilbronn}. These results suggest that the components in ReASearch are not incidental tools: they support different aspects of agent-controlled search, and together let the same scaffold cover diverse optimization tasks.

\begin{table}[h!]
\centering
\caption{Component ablation on prompt optimization.}
\label{tab:ablation_prompt}
\resizebox{0.7\linewidth}{!}{%
\begin{tabular}{lcc}
\hline
Method & AIME $\uparrow$ & Terminal-Bench 2.0 $\uparrow$ \\
\hline
Baseline                   & 46.00 & 35.56 \\
GEPA                       & 50.67 & 42.22 \\
ReASearch                  & \textbf{52.00} & \textbf{53.33} \\
ReASearch w/o memory       & 49.33 & 48.15 \\
ReASearch w/o Python tools & 51.33 & 51.11 \\
\hline
\end{tabular}%
}
\end{table}

\begin{table}[h!]
\centering
\caption{Component ablation on ARC-AGI-2 program evolution.}
\label{tab:ablation_arc}
\resizebox{0.7\linewidth}{!}{%
\begin{tabular}{lcc}
\hline
Method & Train acc $\uparrow$ & Test acc $\uparrow$ \\
\hline
AdaEvolve                  & 21.9\% & 12.5\% \\
ReASearch                  & \textbf{85.0\%} & \textbf{50.0\%} \\
ReASearch w/o memory       & 60.0\% & 39.2\% \\
ReASearch w/o Python tools & 51.67\% & 32.5\% \\
\hline
\end{tabular}%
}
\end{table}

\paragraph{Memory design and transfer.}
We study whether lessons transfer across runs by initializing a new run with the \texttt{lessons.md} file from a NanoGPT run that achieved 0.969 bpb. The new agent automatically updates the baseline and starts at 0.977, rather than the default baseline loss of about 0.998, suggesting that the memory file captures reusable optimization knowledge rather than run-specific traces. We also compare different memory schemas and find that a simple schema---\emph{what worked}, \emph{what did not work}, \emph{what to try next}---is sufficient: across several prompt-optimization experiments on AIME and HotpotQA, it performs similarly to, and sometimes about 1\% better than, richer task-specific memory instructions, which can even hurt performance by encouraging rigid or noisy summaries.

\paragraph{Validation feedback granularity.}
We ablate how much validation information is exposed during prompt optimization. Providing only the aggregate validation metric works better than exposing per-example validation trajectories, which leads to clear overfitting. On AIME, for example, performance drops from 52\% to around 50\% when detailed validation trajectories are revealed.

\section{Open-source backbones}
\label{app:open_source}

ReASearch is not tied to a specific proprietary model: any model with sufficient long-horizon agentic ability can serve as the optimizer, and performance generally scales with the capability of the underlying model. To verify this, we evaluate ReASearch with two strong open-source models, GLM-5 and Kimi-2.5, on prompt optimization (Table~\ref{tab:open_prompt}) and ML workflow optimization (Table~\ref{tab:open_ml}). On AIME and HotpotQA, where these models have base capabilities comparable to Claude Sonnet 4.6, the open-source results closely match Sonnet 4.6. On the more challenging Terminal-Bench 2.0 setting with GPT-OSS-120B as the student model, the baseline is near zero, and ReASearch with GLM-5 and Kimi-2.5 lifts it to roughly 15\% and 8\%, respectively. On ML workflow optimization, open-source backbones are slightly weaker than Claude/GPT-based agents, as expected, but still show clear improvements over the baseline.

\begin{table}[h!]
\centering
\caption{Open-source backbones on prompt optimization. The AIME and HotpotQA rows for the baseline and ReASearch w/ Claude Sonnet 4.6 match Table~\ref{tab:prompt}; the Terminal-Bench 2.0 column here uses GPT-OSS-120B as the student model (rather than GPT-5), so its values differ from Table~\ref{tab:prompt} and are not directly comparable. GLM-5 and Kimi-2.5 are added.}
\label{tab:open_prompt}
\resizebox{\linewidth}{!}{%
\begin{tabular}{lccc}
\hline
Method & AIME $\uparrow$ & HotpotQA $\uparrow$ & Terminal-Bench 2.0 $\uparrow$ \\
\hline
Baseline                       & 46.00 $\pm$ 1.33 & 63.00 $\pm$ 1.50 & 3.00 $\pm$ 0.74 \\
GEPA                           & 50.67 $\pm$ 1.15 & 65.80 $\pm$ 0.80 & --- \\
ReASearch w/ Claude Sonnet 4.6 & \textbf{52.00 $\pm$ 0.67} & \textbf{67.60 $\pm$ 0.50} & \textbf{23.00 $\pm$ 2.27} \\
ReASearch w/ GLM-5             & 51.33 $\pm$ 1.15 & 67.20 $\pm$ 0.90 & 14.81 $\pm$ 4.07 \\
ReASearch w/ Kimi-2.5          & 50.67 $\pm$ 0.94 & 66.80 $\pm$ 1.00 & 8.15 $\pm$ 3.39 \\
\hline
\end{tabular}%
}
\end{table}

\begin{table}[h!]
\centering
\caption{Open-source backbones on ML workflow optimization. The baseline and ReASearch w/ Claude rows match Table~\ref{tab:ml}; GLM-5 and Kimi-2.5 are added.}
\label{tab:open_ml}
\resizebox{\linewidth}{!}{%
\begin{tabular}{lccccc}
\hline
Method & NanoGPT $\downarrow$ & IMG-100 $\uparrow$ & Atari (Q*bert) $\uparrow$ & MuJoCo $\uparrow$ & Crypto $\uparrow$ \\
\hline
Baseline              & 0.998 $\pm$ 0.006 & 63.51 $\pm$ 0.85 & 475 $\pm$ 90 & 1537 $\pm$ 220 & 0.0953 $\pm$ 0.0021 \\
ReASearch w/ Claude   & \textbf{0.976 $\pm$ 0.008} & \textbf{83.99 $\pm$ 1.10} & \textbf{4500 $\pm$ 320} & \textbf{5267 $\pm$ 480} & \textbf{0.1110 $\pm$ 0.0024} \\
ReASearch w/ GLM-5    & 0.979 $\pm$ 0.009 & 82.80 $\pm$ 1.25 & 4100 $\pm$ 380 & 4920 $\pm$ 510 & 0.1085 $\pm$ 0.0030 \\
ReASearch w/ Kimi-2.5 & 0.981 $\pm$ 0.010 & 82.10 $\pm$ 1.35 & 3900 $\pm$ 420 & 4740 $\pm$ 540 & 0.1076 $\pm$ 0.0033 \\
\hline
\end{tabular}%
}
\end{table}

\section{Component ablation on Heilbronn triangle}
\label{app:ablation_heilbronn}

We extend the component ablation in Section~\ref{sec:experiments} to the Heilbronn triangle program-evolution task. As in the ARC-AGI-2 ablation, removing the Python execution tool has a slightly larger effect than removing memory, consistent with the program-evolution setting where short-horizon analysis and verification are central.

\begin{table}[h!]
\centering
\caption{Component ablation on Heilbronn triangle program evolution. The objective is the minimum triangle area; higher is better.}
\resizebox{\linewidth}{!}{%
\begin{tabular}{lccccc}
\hline
Method & $n=11$ $\uparrow$ & $n=12$ $\uparrow$ & $n=13$ $\uparrow$ & $n=14$ $\uparrow$ & $n=15$ $\uparrow$ \\
\hline
Best known (human)         & 0.03704 & 0.03260 & 0.02702 & 0.02430 & 0.02111 \\
AdaEvolve                  & 0.02629 & 0.02552 & 0.02142 & 0.00299 & 0.00984 \\
ReASearch                  & \textbf{0.03552} & \textbf{0.03260} & \textbf{0.02700} & \textbf{0.02429} & \textbf{0.02034} \\
ReASearch w/o memory       & 0.03241 & 0.02938 & 0.02371 & 0.02064 & 0.01617 \\
ReASearch w/o Python tools & 0.03387 & 0.02862 & 0.02497 & 0.01892 & 0.01753 \\
\hline
\end{tabular}%
}
\end{table}

\section{Claude Code baseline details and comparison analysis}
\label{app:claude_code}

\paragraph{Baseline configuration.}
The Claude Code baseline is the official setup in AutoResearch\footnote{https://github.com/karpathy/autoresearch}. It is the standard Claude Code CLI with its full default tool surface, initialized with the upstream instruction file verbatim. That instruction file already prescribes the optimization goal, the only-edit-the-training-script rule, the per-experiment training-time budget, the keep-on-improvement / discard-on-regression loop, the simplicity criterion, and the never-stop rule. The backbone LLM is Claude Sonnet 4.6, matching ReASearch on all reported runs. The training script, data pipeline, evaluator, hardware, and per-experiment training cap are identical on both sides.

\paragraph{Matched instructions, different harness.}
The instructions on the two sides say the same things. Claude Code reads Karpathy's \texttt{program.md} (goal, only-edit-\texttt{train.py} rule, time budget, keep/discard loop, simplicity, NEVER-STOP, crash/VRAM guidance). ReASearch's ML-workflow system prompt states the same goal, the same edit-\texttt{train.py} rule, the same time budget, the same loop (``LOOP FOREVER: come up with idea $\rightarrow$ edit $\rightarrow$ commit $\rightarrow$ run $\rightarrow$ keep-or-reset''), the same simplicity and NEVER-STOP rules. The phrasing differs; the content is matched. The score gap therefore does not come from ``ReASearch was told something Claude Code wasn't.''

\paragraph{Where the gain comes from: re-emission, not extra rules.}
The architectural difference is not simply that ReASearch has more instructions. The difference is \emph{where} those instructions live and how often they are surfaced. Claude Code is a skill-based, black-box-heavy agent system: the system prompt is fixed at conversation start, while additional information is exposed through agent-invocable surfaces such as \texttt{Read}/\texttt{Write}/\texttt{Bash}, MCP servers, skills, hooks, and an auto-memory file. The agent decides when to consult those surfaces; what it does not invoke, it does not see.

ReASearch instead splits the same content into two layers. A small static layer contains general rules and the search-loop description. The remaining state is \emph{re-emitted into the system prompt every turn}: the current best result, recent experiments, persistent lessons, stagnation warnings, and a structured view of the search tree. The gain comes from making search state unavoidable, rather than merely available.

Concretely, five behaviors do the load-bearing work:
\begin{enumerate}
\item \textbf{Goal and metric re-rendering.} The goal, metric, and optimization direction are shown every turn, so the agent does not need to re-derive them from a file.

\item \textbf{Recent experiment memory.} The last 20 experiments are rendered as a structured table with parent, metric, memory, status, and description. This addresses long-horizon failures where agents re-try variants they had already invalidated, as in 1{,}800+-turn IMG-100 runs.

\item \textbf{Persistent lessons.} \texttt{lessons.md} is carried through context compaction, preserving the \emph{why} of past failures, such as ``doubling envs didn't help---bottleneck is FC width'' or ``PER is 133$\times$ slower on this hardware.''

\item \textbf{Unconditional stagnation advice.} After 3+ and 7+ flat evaluations, a stagnation advisory is inserted directly into the prompt. This matters because an agent on a plateau is unlikely to realize that it should call a diagnostic tool.

\item \textbf{Structured search-tree review.} \texttt{review\_search\_tree} turns flat history into branch concentration, family distribution, and plateau signals, so the agent reads search history as structure rather than as TSV rows.

\end{enumerate}

This is why skills and MCP servers are not enough by themselves. They run only when the agent thinks to invoke them, which is exactly what often fails on plateaus and over long horizons. Since Claude Code’s system prompt is fixed for the conversation, per-turn re-rendering is not a configuration knob. Obtaining this behavior from a default Claude Code setup requires rebuilding the outer loop, not merely enabling more skills.

More precisely, this is not a claim that Claude Code's runtime cannot host such behaviors. Its hook mechanism could, in principle, maintain a stateful ledger, parse tool outputs, and re-emit a structured experiment table after every turn. But realizing this through hooks layered on a black-box agent is closer to engineering a new framework than to configuring an existing one, and the result is the same behavior reached through a heavier and less controllable route. ReASearch can be viewed as the principled form of that path: a first-class outer loop that owns prompt assembly directly, with a persistent Python namespace replacing per-call shell one-shots.

\section{ReASearch details}
\label{app:detail}

In this section, we provide the full tool specifications exposed to the agent in our scaffold. All tasks share a common set of general-purpose tools (Section~\ref{app:general_tools}), and each task family adds domain-specific tools (Section~\ref{app:specific_tools}).

\subsection{General tools}
\label{app:general_tools}

The following tools are available to the agent across all task types.

\paragraph{File I/O.}
The agent can read and write files within its workspace directory.
\begin{itemize}[leftmargin=*,itemsep=2pt]
\item \texttt{read\_file(path)}: 
Paths are sandboxed to the workspace; directory traversal is blocked. Reading validation evaluation logs is blocked to prevent overfitting.
\item \texttt{write\_file(path, content)}: 
Creates parent directories as needed.
\item \texttt{list\_files(path)}: 
Returns file names, types, and sizes.
\end{itemize}

\paragraph{Code execution.}
\begin{itemize}[leftmargin=*,itemsep=2pt]
\item \texttt{python\_exec(code)}: \
Each execution is subject to a configurable timeout (default 300\,s) and is logged to disk.
\item \texttt{bash(command)}: 
An extensive blocklist prevents destructive operations (e.g., \texttt{rm -rf}, \texttt{chmod 777}, pipe-to-shell patterns). 
\end{itemize}

\paragraph{Context management and memory.}
The agent operates under a finite context window. When the conversation exceeds a token threshold (default 90{,}000 tokens), the scaffold automatically compresses the context. The agent can also trigger this manually:
\begin{itemize}[leftmargin=*,itemsep=2pt]
\item \texttt{compact()}: 
The scaffold saves the full transcript to disk, calls the agent LLM to produce a summary of key insights and unfinished work, then replaces the entire message history with a compact 3-message sequence: an authoritative state snapshot (budget, scores, candidates), the conversation summary, and the contents of \texttt{lessons.md}.
\item \texttt{get\_status()}: 
Each task type overrides this with domain-specific state.
\end{itemize}

The agent also maintains a \texttt{lessons.md} file in its workspace, which serves as persistent memory across context compressions. The system prompt instructs the agent to record key learnings (what worked, what didn't, what to try next) to this file, and its contents are automatically injected into the compressed context whenever compression occurs. When launching a new run, lessons from a previous run can be pre-loaded via a \texttt{{-}{-}lessons} flag, enabling cross-run transfer of insights.

\subsection{Domain-specific tools}
\label{app:specific_tools}

\subsubsection{Prompt optimization}

In prompt optimization, the agent optimizes a textual artifact (e.g., a system prompt) for a student language model. The agent is given the following tools in addition to the general tools. Each tool call consumes budget equal to the number of student-model calls it makes.

\begin{itemize}[leftmargin=*,itemsep=2pt]
\item \texttt{get\_dataset\_info()}: 
Get information about a dataset: size, sample problem, expected answer.

\item \texttt{get\_next\_minibatch(size)}: 
Call the sampler to draw a specified number of examples from the training set and return the indices.

\item \texttt{call\_student\_model\_batch(prompt\_text, sample\_indices)}: 
This allows the agent to test prompt drafts on customized small training subsets before committing to full validation.

\item \texttt{validate\_candidate(prompt\_text)}: 
Run evaluation on the validation set and receive only the aggregate metric, without access to the detailed trajectories.

\end{itemize}

\subsubsection{Program evolution}

In program evolution, the agent iteratively modifies a Python program to optimize a task-specific score. The following domain-specific tools are provided:

\begin{itemize}[leftmargin=*,itemsep=2pt]
\item \texttt{get\_current\_code()}: 
Read the current code from \texttt{solve.py}.

\item \texttt{edit\_code(instruction)}: 
A separate LLM call reads the current code and the instruction, then outputs the full modified file. This allows the agent to describe changes concisely rather than producing entire programs inline.

\item \texttt{evaluate(code\_text?)}: 
Evaluate the current code against the problem and get score.

\end{itemize}

\subsubsection{ML workflow optimization}

In ML workflow optimization, the agent modifies a training script and runs experiments. The following domain-specific tools are provided:

\begin{itemize}[leftmargin=*,itemsep=2pt]

\item \texttt{get\_current\_code()}: 
Read the current code from \texttt{train.py}.

\item \texttt{edit\_train\_file(instruction)}: 
A separate LLM call reads the current training code and the instruction,
then outputs the full modified file. This allows the agent to describe changes concisely
rather than producing entire programs inline.

\item \texttt{run\_experiment(description)}: 
Run a time-bounded training job, and return the related metrics.

\item \texttt{git(command)}: 
Run a git command in the task repo directory. Use for: git commit, git log, git diff, git status, git reset, git checkout. Push and remote operations are blocked.

\end{itemize}

\section{Experiment settings}
\label{app:setting}
\subsection{Prompt optimization}
\paragraph{AIME}
We use 45 problems from AIME 2022, 2023, and 2024 for training and the remaining 45 for validation, repeating each validation evaluation 5 times to reduce variance. We use AIME 2025 (30 problems) as the test set and evaluate performance both before and after prompt optimization, again averaging over 5 runs. We use GPT-4.1 mini as the student model with temperature 1.0 and set the student evaluation budget to 1000.

\paragraph{HotpotQA}
We use the HotpotQA dataset in the distractor setting \citep{yang2018hotpotqa}, drawn from the validation split (7,405 examples). After shuffling with a fixed seed, we use the first 500 examples for training, the next 500 for validation, and 1,000 for testing. Performance is measured using exact match after normalizing case, articles, and punctuation. We use GPT-4.1 mini as the student model with temperature 1.0 and set the student evaluation budget to 5000.

\paragraph{GSM8K}
We use GSM8K \citep{gsm}, taking 500 shuffled examples from the training split for optimization and another 500 for validation. The full official test split (1,319 examples) is used for testing. We use Llama 3.1 as the student model with temperature 0.7 and set the student evaluation budget to 5000.

\paragraph{Terminal-Bench}
We use Terminal-Bench 2.0 \citep{merrill2026terminal}, which comprises 89 tasks. We assign 22 tasks to training, 22 to validation, and the remaining 45 to testing, repeating both validation and test evaluations 3 times to reduce variance. Following the Terminus 2 setup, the student agent is limited to a headless terminal as the single tool, and interacts with the environment exclusively through Bash commands. Tasks are evaluated via Docker-based verification with binary pass/fail outcomes. We use GPT-5 as the student model and set the student evaluation budget to 300.

\subsection{Program evolution}
\paragraph{Circle packing.}
We benchmark the max-sum variant where we place $n$ disjoint circles into a unit square and maximize the sum of radii. We test $n$ from 23 to 32 (10 instances) with best known results from \citet{friedmanpacking}. The agent starts from a simple grid-layout baseline and evolves the packing strategy. Each evaluation runs with a 600-second timeout. We report the achieved sum of radii. The maximum budget is set to 500.

\paragraph{Heilbronn triangle.}
We place $n$ points in a unit square to maximize the minimum triangle area over all $\binom{n}{3}$ triples. We test $n$ from 11 to 15 (5 instances) with the best known results from \citep{heilbronn}. The agent starts from a stochastic local search baseline that perturbs points from a jittered grid. Each evaluation runs with a 300-second timeout. We report the achieved minimum triangle area. The maximum budget is set to 500.

\paragraph{EPLB.}
We optimize the expert-parallelism load balancer for a Mixture-of-Experts model, using a real load trace with 288 expert replicas across 32 GPUs and 4 nodes. The agent starts from the published DeepSeek EPLB algorithm \citep{deepseekeplb} and evolves the rebalancing strategy. Each evaluation simulates inference across sliding windows and runs with a 360-second timeout. We report a composite score: $(b_{\text{expert}} + s_{\text{speed}}) / 2$, where $b_{\text{expert}}$ measures load balancedness and $s_{\text{speed}} = 0.002 / t_{\text{inference}}$. The maximum budget is set to 500.

\paragraph{Transaction scheduling.}
We schedule 3 workloads of 100 database transactions each to minimize the total makespan, subject to read/write conflict dependencies. The agent starts from a greedy cost-sampled scheduler baseline and evolves the scheduling strategy. Each evaluation runs with a 600-second timeout. We report the total makespan across all three workloads. The maximum budget is set to 500.

\paragraph{ARC-AGI-2.}
We use 120 puzzles from the ARC-AGI-2 evaluation set \citep{chollet2025arc}. For each puzzle, the agent writes two transformation functions (\texttt{transform\_grid\_attempt\_1} and \texttt{transform\_grid\_attempt\_2}) that map input grids to output grids based on a few training examples. The agent starts from a dummy placeholder baseline (rotation + color increment) that must be completely rewritten. Each evaluation runs both attempts with a 60-second timeout. We report accuracy (pass@2): a puzzle is solved if either attempt exactly matches the expected output. The maximum budget is set to 500.

\subsection{ML workflow optimization}

\paragraph{NanoGPT.}
The agent improves a GPT language model in the NanoGPT setting \citep{nanogpt}, minimizing validation bits per byte (\texttt{val\_bpb}). The time budget is 5 minutes with a 10-minute experiment timeout.

\paragraph{IMG-100.}
The agent improves image classification on IMG-100, maximizing accuracy on the validation set. The baseline is a ResNet-18. The time budget is 5 minutes with a 10-minute experiment timeout.

\paragraph{Atari.}
The agent improves deep RL policies for the Q*bert environment, maximizing evaluation reward averaged over 10 episodes. The baseline is PPO with a 30-minute time budget and 40-minute experiment timeout.

\paragraph{MuJoCo.}
The agent improves continuous control policies on HalfCheetah, maximizing evaluation reward averaged over 10 episodes. All environments use SAC as the baseline algorithm. The time budget is 10 minutes with a 15-minute experiment timeout.

\paragraph{DRW - Crypto Market Prediction}\footnote{https://www.kaggle.com/competitions/drw-crypto-market-prediction}
The agent optimizes models for an anonymized crypto trading dataset with 785 features and approximately 525K training examples, using validation Pearson correlation (\texttt{val\_corr}) as the optimization objective. We begin from a 6-fold cross-validation baseline that ensembles a linear model and an MLP, whose submission ranks 36th on the Kaggle leaderboard. We then perform a small amount of manual iteration and leaderboard submission, distill the resulting lessons into the agent instructions, and allow the agent to continue optimizing from there. This setup is intended to mimic a realistic Kaggle competition workflow.

\section{Optimization artifacts and agent lessons}
\label{app:artifacts}

We present the full prompts (before and after optimization) for each prompt optimization task for both ReASearch and GEPA, and the summarized agent's self-written \texttt{lessons.md} files across all task categories. These lessons are maintained by the agent throughout each run and survive context compression; they serve as the agent's persistent memory of what worked, what didn't, and what to try next.

\subsection{Prompt optimization}

\subsubsection{AIME}

\paragraph{Baseline prompt.}
\mbox{}\\\begin{lstlisting}[style=prompt]
Solve the problem and provide the answer in the correct format.

Reasoning: Let's think step by step in order to solve this problem.

After your reasoning, output your final answer as ### N where N is an integer (0-999).
\end{lstlisting}

\paragraph{ReASearch optimized prompt.}
\mbox{}\\\begin{lstlisting}[style=prompt]
You are solving an AIME-style math problem. Compute a single integer N with 0 <= N <= 999, then output exactly one final line: ### N

Method (concise and exact):
- Clarify the target: Restate exactly what is asked, for example reduced m/n then m+n; least nonnegative residue; total number of labeled configurations; a length or area leading to m*sqrt(n) then m+n. Re-read the last sentence before finalizing.
- Exact math only: use integers, fractions, and radicals with square-free parts; rationalize when needed. No decimal approximations unless rounding is requested. Do not cite "known answers" or use code.
- Track constraints: ordered vs unordered; labeled vs unlabeled, and count all labeled configurations unless symmetry reduction is explicitly required; distinctness; parity; digit/base bounds; coprimality; domains.
- AIME toolkit: algebra/factorization; Vieta/discriminants; modular arithmetic; complementary counting/inclusion-exclusion; invariants/symmetry; exact geometry including similarity, Power of a Point, Ptolemy, mass points, and clean coordinates.
- Patterns: If p(m)=p(k), write p(x)-p(k)=(x-k)r(x); analyze r with Vieta/discriminant and enforce integer/uniqueness by bounded casework.
- Prefer finite casework when ranges are small, such as digits or small divisors; set up explicit constraints rather than guessing.
- Validate any derived formula on at least two simple cases, including edge or given examples; if a test fails, fix the derivation.

Final checklist:
1) Target match: reduced m/n before m+n; least nonnegative residue; square-free radicals before m+n; count labeled objects unless told otherwise; correct object and complete case coverage.
2) N is a single integer in [0,999].
3) Sanity checks: parity/mod and magnitude bounds; quick plug-back when feasible.
4) Output exactly one line: ### N with no extra text after.
\end{lstlisting}

\paragraph{GEPA optimized prompt.}
\mbox{}\\\begin{lstlisting}[style=prompt]
Task: You will be given a single contest-style math problem. Solve it exactly and output ONLY the final numeric answer as a bare integer with no extra text, symbols, or formatting.

Output format (strict):
- Output only the final integer.
- No words, labels, LaTeX, boxes, units, punctuation, or trailing/leading zeros.
- Examples of valid outputs: 756, 751, 73.
- Examples of invalid outputs: \boxed{756}, 073, 756., Answer: 756.

Core directives:
- Read the prompt carefully to identify exactly what quantity is requested (e.g., AC^2 instead of AC; or m+n for a fraction m/n in lowest terms).
- Keep all computations exact (no decimal rounding). Use radicals, fractions, and exact integers as needed until the final step.
- If the problem defines a fraction m/n with m and n relatively prime, fully reduce the fraction before computing m+n.
- If a squared length is requested, compute it directly (avoid unnecessary square roots).
- Perform quick consistency checks for reasonableness before finalizing.
- Use bounding/degree checks to prune impossible contributions in sums/series/polynomials.
- If the problem asks for a sum of squares or similar aggregate, compute exactly and only at the end.
- If multiple cases are possible, ensure the answer satisfies all given conditions; eliminate extraneous solutions.

General problem-solving guidance:
- Introduce substitution to simplify structure (e.g., set x = log_b n, or parameterize z = r e^{i\theta}).
- Use symmetry, parity, and monotonicity to reduce cases.
- For constrained optimization on circles/spheres, convert to linear forms and use Cauchy--Schwarz or a distance-to-line interpretation.
- For medians and modes in integer lists, relate list length parity to median definition; enforce uniqueness conditions on frequencies.

Domain-specific strategies and facts:

1) Geometry with spheres and a cutting plane:
- For spheres of radii r_i with centers O_i, intersected by a plane in congruent circles of radius x with all centers on the same side of the plane:
  x^2 = r_i^2 - h_i^2 with h_i > 0 where h_i is the perpendicular distance from O_i to the plane.
- If A, B are the projections of O_i, O_j onto the plane (centers of the intersection circles), then:
  AB^2 = |O_iO_j|^2 - (h_i - h_j)^2.
- If spheres are mutually externally tangent, then |O_iO_j| = r_i + r_j.
- Use given planar distances (e.g., AB^2) to solve for x^2 and h_i; then compute any requested planar distances via the same AB^2 relation.

2) Cube geometry with tilted orientation and horizontal water surface:
- A vertical face can be used to model linear variation of vertex heights above the horizontal plane; heights vary linearly along edges.
- Determine the cube side length s from given vertex heights using orthogonality or similar triangles; in a known analogous setup, s = 6.
- Volume with a horizontal plane at height h: the empty region above the water may be a frustum between two parallel planar sections. Use:
  V_frustum = (H/3) * (A1 + sqrt(A1*A2) + A2),
  where H is the separation along the normal inside the solid, and A1, A2 are the cross-sectional areas at the two levels.
- Practical approach: compute the full cube volume s^3; compute the empty frustum volume by identifying cross-sectional squares that scale linearly; subtract from s^3.

3) Set counting with "everyone owns" one item and exact-count constraints:
- If all N people own a universal item H, and other items A, B, C have totals |A|, |B|, |C|; let w, x, y, z be the counts owning exactly 1, 2, 3, 4 items (including H).
- Then:
  w + x + y + z = N,
  w + 2x + 3y + 4z = T where T = |A| + |B| + |C| + |H| and |H| = N.
- If some of w, x, y, z are given, solve the linear system to find the desired count (e.g., z).

4) Rectangles in a regular 2n-gon (e.g., dodecagon):
- Rectangles correspond to choosing two distinct pairs of opposite vertices. In a regular dodecagon, there are 6 opposite pairs; each rectangle is determined by two of these pairs.
- Model as a 2x6 grid by pairing opposite vertices as columns; rectangles correspond to 2x2 submatrices formed by two columns.
- For 2-colorings avoiding a monochromatic rectangle:
  - Column types are (top,bottom) in {(0,0), (0,1), (1,0), (1,1)}.
  - Types (0,0) and (1,1) may appear at most once each; (0,1) and (1,0) may repeat freely.
  - Casework for a dodecagon yields a total of 928 valid colorings.

5) Coefficients in generating functions of the form P(x) = (x^N - 1)^k / \prod(x^{m_i} - 1) for 0 < x < 1:
- Use 1/(x^m - 1) = -1/(1 - x^m) = -\sum_{t>=0} x^{mt} and expand (x^N - 1)^k = \sum_{j=0}^k (-1)^{k-j} C(k, j) x^{Nj}.
- If the target exponent n satisfies n < N, then only j = 0 contributes.
- Reduce to counting nonnegative integer solutions of \sum m_i t_i = n via modular reductions and L = lcm(m_i); often becomes a stars-and-bars count for the multiple-of-L part after fixing residues.
- Example: With m_i in {105,70,42,30} and n = 2022 (< 2310), count solutions to 105b + 70c + 42d + 30e = 2022; successive modular reductions lead to b' + c' + d' + e' = 9, giving C(12,3) = 220.

6) River-crossing with current and equidistant target on opposite bank:
- Let x be downstream and y across the river; current adds (v_c, 0) to each swimmer's water-relative velocity.
- If swimmers start at (0,0) and (D,0) and meet simultaneously at (D/2, W) in time T:
  - Vertical components satisfy v_y = W/T.
  - Ground-velocity x-components include current; magnitude constraints yield two equations in unknowns.
  - Subtract to eliminate squares, giving a linear relation D = c T for constant c; then use W = v_y T to solve for T and compute D.

7) Complex numbers: maximizing the real part with |z| fixed:
- For expressions of the form Re(A z + B/z) with |z| = r:
  - Use 1/z = conjugate(z)/|z|^2 = (1/r) e^{-i\theta} if z = r e^{i\theta}.
  - Then Re(A r e^{i\theta} + (B/r) e^{-i\theta}) becomes a linear combination p cos \theta + q sin \theta; the maximum is sqrt(p^2 + q^2).
  - Equivalent approach: set z = a + bi with a^2 + b^2 = r^2; express the real part as P a + Q b and maximize via Cauchy--Schwarz: max = r sqrt(P^2 + Q^2).
  - Tangent-to-circle distance method gives the same maximum.

8) Integer lists with sum, unique mode, and "median is an integer not in the list":
- If the median is an integer but not in the list, the list length must be even. The median equals (a_m + a_{m+1})/2, an integer that must not appear as any element.
- The two middle elements must have the same parity (their average is an integer). A common minimal pattern is they differ by 2 so the average is the in-between integer not present.
- Unique mode "9" means the count of 9's is strictly greater than the count of any other value; avoid ties (e.g., do not allow another value to occur as often as 9).
- For small total sum constraints, start with the smallest even length that can realize the requirements (often n=4). With sum 30 and unique mode 9:
  - Two 9's force the remaining two numbers to sum to 12 and be less than 9 to prevent median 9.
  - To make the median an integer not in the list, choose the two to be (5,7) (not (6,6), which breaks uniqueness of the mode).
  - Then compute the requested aggregate (e.g., sum of squares 5^2 + 7^2 + 9^2 + 9^2 = 236).

9) Logarithmic equations with base b != 1:
- Let x = log_b n to linearize expressions. Use log_b(n^k) = k log_b n and log_b(bn) = 1 + log_b n.
- Respect domains: sqrt(log_b n) requires log_b n >= 0.
- Example: sqrt(x) = x/2 and b x = 1 + x => x in {0,4}, discard x=0 if it breaks the second equation; with x=4 obtain b = (1+x)/x = 5/4 and n = b^x = (5/4)^4 = 625/256; reduce before computing j+k.

Process checks:
- Confirm all constraints (parity, domain, uniqueness, ranges) are satisfied by the final candidate.
- Discard extraneous roots introduced by squaring or by implicit domain violations.
- For expressions requiring a maximum/minimum, justify that the found value is achievable.

Final step:
- After verifying correctness, output only the final integer as specified above.
\end{lstlisting}

\paragraph{Agent lessons.}
\mbox{}\\\begin{lstlisting}[style=lesson]
## What Worked
- Baseline system prompt (#1) achieved 51.1% on val (mean over 5 runs). On train, it scored 27/45 = 60%. Formatting was generally correct.
- Cand5_prompt (Candidate #3) performed best on train: 66.7%. Key elements:
  - exact arithmetic (no decimals),
  - target clarity and strict final checklist,
  - pattern guidance for p(m)=p(k) via factoring and Vieta/discriminant,
  - finite casework for small ranges.

## What Didn't Work
- Cand7_prompt added early parity/mod checks and Euler/arrangement guidance, but validation dropped to 45.8% -- overloading/overfitting hurt generalization.
- Cand2_prompt regressed to 48.9%, with hallucinated "known answers," decimal approximations, and out-of-range outputs.
- Cand8_prompt tried to hard-code cubic uniqueness, but train slipped to 53.3%. Strong assertions may be wrong in some cases or too prescriptive.

## Key Lessons
- The baseline is already near-optimal for AIME with this student model.
- Domain-specific add-ons consistently degraded val performance.
- Adding rules that help specific train problems tends to hurt generalization.
\end{lstlisting}

\subsubsection{HotpotQA}

\paragraph{Baseline prompt.}
\mbox{}\\\begin{lstlisting}[style=prompt]
You are a question answering assistant. You will be given context documents and a question. Read the documents carefully and answer the question.

Output your final answer as ### <answer> on its own line.
\end{lstlisting}

\paragraph{ReASearch optimized prompt.}
\mbox{}\\\begin{lstlisting}[style=prompt]
You are an expert question answering assistant for multi-hop reasoning tasks. You will be given several context documents and a question that requires reasoning across multiple documents.

## Your Task
Read all documents carefully, then answer the question by following the chain of reasoning across documents.

## Reasoning Process

**Step 1 -- Build the reasoning chain**: Multi-hop questions require connecting facts across multiple documents. Trace each step: what does document A tell you? How does that connect to document B? Follow the chain to the FINAL answer.

**Step 2 -- Determine the answer type from the question**:
- "Who..." -> a person's name (use the full name as it appears in the document, including middle names)
- "Which [entity]..." -> the name of that specific entity
- "How many..." -> the number or quantity
- "What [year/date]..." -> the year or date
- "What [title/name/role/type/kind]..." -> the exact title, name, role, or type
- "Is/Are/Was/Were/Did/Does/Can..." -> "yes" or "no" only

**Step 3 -- Extract the answer span**: Find the exact phrase in the documents that answers the question. The answer should be:
- A direct quote from the document (verbatim)
- The MINIMUM phrase that fully answers the question
- Not a full sentence -- just the answer itself
- Not surrounded by extra context words

**Step 4 -- Verify your answer**:
- Does it directly answer what the question asks?
- Is it the right type? Check carefully:
  - If asked "who" -> must be a person's name, not an organization or place
  - If asked "what kind/type/genre" -> must be a category, not a specific instance
  - If asked "which [film/book/team]" -> must be the name of that thing
  - If asked "what year" -> must be a year (or date)
- Have you stopped at the right point -- not too much, not too little?

## Key Rules
- **Yes/No questions**: Answer ONLY "yes" or "no" -- nothing else.
- **Comparison questions** (who is older/bigger/has more X): give the NAME of the winning entity, not the value.
- **Full names**: Use the complete name as it appears in the document, including middle names.
- **No extra context**: Do not add "by [author]", "in [place]", or other surrounding words.
- **No paraphrase**: Copy the exact wording from the document.
- **Don't drop words**: Keep all words that are part of the answer.

## Format
Think through the reasoning step by step, then output your final answer on its own line as:
### <answer>

The answer after ### should contain ONLY the answer itself.
\end{lstlisting}

\paragraph{GEPA optimized prompt.}
\mbox{}\\\begin{lstlisting}[style=prompt]
You are a question answering assistant. You will be given context documents and a question. Read the documents carefully and answer the question. Give a short, concise answer (typically 1-3 words).

Important guidelines:
1. Base your answer strictly on the information provided in the documents.
2. When the question asks about a specific detail, find the exact phrasing used in the relevant document and use that as your answer.
3. Pay close attention to the exact wording in the documents. Use the most specific or complete form of the answer as it appears in the text (e.g., full names, exact titles, precise terms).
4. If a person's full name is given in the document (e.g., "Masayoshi 'Masa' Takayama"), use the complete name as written, not a shortened version.
5. If a disease or term has multiple names in the document (e.g., "amyotrophic lateral sclerosis (ALS), commonly known as 'Lou Gehrig's Disease'"), use the name that best matches the phrasing of the question or the most commonly referenced form in the document.
6. Do not paraphrase, summarize, or infer beyond what is explicitly stated in the documents.
7. If the question references a nickname or alternate name, map it to the full name as given in the document.
\end{lstlisting}

\paragraph{Agent lessons.}
\mbox{}\\\begin{lstlisting}[style=lesson]
## Improvement Chain
- Baseline (#1): 8.8%
- V2 (#2): 21.2% -- structured approach + yes/no instruction
- V4 (#4): 60.0% -- question-type awareness, verbatim extraction, full names
- V5 (#5): 65.0% -- question-type guidance (who->person, which->entity)
- V10 (#10): 66.0% -- "MINIMUM phrase" framing, verify step
- V14 (#14): 68.0% -- explicit type-mismatch checks in verify step

## What Made #14 the Best
Compared to #10 (66%), #14 added to the verify step:
- "If asked 'who' -> must be a person's name, not an organization or place"
- "If asked 'what kind/type/genre' -> must be a category, not a specific instance"
These explicit type-mismatch checks pushed from 66% to 68%.

## What Hurt (every modification to #14 degraded val)
- Removing "use full names" rule: 65% -> 62.8%
- Adding stronger anti-verbosity rules: 65% -> 61.2%
- Adding worked examples: 65% -> 62.2%
- Shorter/simpler prompt: 66% -> 57% (MAJOR REGRESSION)
- Adding "how many -> number only without units": 68% -> 66.8%
- Adding "don't repeat question nouns": 68% -> 65.6%
- Adding "Re-read question": 68% -> 66.0%

## Key Insights
- The detailed 4-step structure is CRITICAL -- shorter prompts hurt significantly.
- Train batch scores are VERY NOISY -- #17 showed 70% vs 65% on train but 66% vs 68% on val.
- Adding rules tends to hurt: every time we add specific rules, they help some cases but hurt others.
- Dataset is inconsistent: sometimes needs full name, sometimes common name.
- We are at a local optimum with #14: every modification tried has hurt on val.
\end{lstlisting}

\subsubsection{GSM8K}

\paragraph{Baseline prompt.}
\mbox{}\\\begin{lstlisting}[style=prompt]
Solve the math problem step by step. Show your work clearly.

After your reasoning, output your final numerical answer as ### N where N is the number (e.g., ### 42).
\end{lstlisting}

\paragraph{ReASearch optimized prompt.}\mbox{}\\
\begin{lstlisting}[style=prompt]
You are an expert math tutor solving grade-school math word problems. Solve each problem carefully and accurately.

## Your Approach:
1. **Read carefully**: Identify all given quantities and what the question is asking for.
2. **Plan**: Break the problem into clear steps.
3. **Solve**: Work through each step, showing calculations.
4. **Verify**: Check that your answer makes sense. Ask: Is the answer positive when it should be? Is it in the right direction (e.g., "lost by N" means subtract, not add)? If you get a different result when re-checking, carefully identify the specific error before changing your answer.

## Common Pitfalls to Avoid:
- "X times more than Y" means X*Y more, so the total is Y + X*Y = (X+1)*Y
- "X times fewer/less" means divide by X (e.g., "3 times fewer" = 1/3 of the amount)
- "cuts X% off" means the remainder is (100-X)% of the original
- "90% less" means 10% of the original remains
- Rate problems: if N workers take T time, then 2N workers take T/2 time (inverse relationship)
- Percentages applied sequentially: apply each to the running total, not the original
- Re-read the question: make sure you're answering what was actually asked

## Output Format:
After your solution, you MUST end your response with your final numerical answer on its own line in this exact format:
### N
where N is the number (e.g., ### 42 or ### 3.5)

Do not skip this final line. Always end with ### N.
\end{lstlisting}

\paragraph{GEPA optimized prompt.}\mbox{}\\
\begin{lstlisting}[style=prompt]
Solve the math word problem step by step. Show your work clearly at each step.

Follow these guidelines:
1. Read the problem carefully and identify all the quantities, relationships, and operations needed.
2. Break the problem into sequential steps, performing one operation at a time.
3. At each step, clearly state what you are calculating and why.
4. Be careful to track which quantities apply to which groups (e.g., if a population grows, the new total becomes the base for the next calculation).
5. Watch for multi-stage problems where results from one stage feed into the next stage - make sure to use the updated totals, not the original starting values.
6. Pay close attention to language like "each" - when it says each member of a group does something, multiply by the CURRENT total count of that group, not just the original count.
7. For problems involving sequences or arithmetic progressions, set up an algebraic equation and solve systematically.
8. For age problems: carefully identify the time reference (e.g., "two years ago", "in five years"). Work out each person's age at the specified point in time, apply any given relationships (e.g., "twice as old") at THAT point in time, and then adjust forward or backward to find the present-day age. Do NOT mix up ages from different points in time.
9. For percentage problems: calculate the percentage of the given base value, then apply the result (add or subtract) as directed.
10. Double-check your arithmetic at each step before moving on.
11. After completing all steps, state your final answer clearly.

After your reasoning, output your final numerical answer as ### N where N is the number (e.g., ### 42).
\end{lstlisting}

\paragraph{Agent lessons.}
\mbox{}\\\begin{lstlisting}[style=lesson]
## Improvement Chain
- Baseline (#1): 88.4%
- V2 (#2): 90.2% (+1.8%) -- structured 4-step + pitfalls + ## headers
- V2-sanity (#9): 90.4% (+0.2%) -- Verify step direction check
- V9-language (#10): 90.6% (+0.2%) -- "X times fewer/less = divide by X" pitfall
- V10-verify-trust (#21): 90.8% (+0.2%) -- "if re-check gives different result, identify specific error before changing answer"

## Verify Step Wording Matters Critically
- #21: "If you get a different result when re-checking, carefully identify the specific error before changing your answer" -> 90.8%
- #22: "Trust your step-by-step work -- only change your answer if you can point to a specific error" -> 88.0%
- The difference: #21 says "identify the error" (implies look for it), #22 says "trust your work" (implies don't change). "Trust your work" is too strong -- prevents legitimate corrections.

## "Pitfall Saturation" Effect -- CONFIRMED
After 7 pitfall bullets, adding ANY more consistently hurts (13+ attempts, all worse). The model has limited attention for instructions; more pitfalls cause confusion. The sweet spot is exactly the 7 pitfalls in #21.

## Key Failure Mode Discovered
- Model correctly computed 40 (Alia markers problem)
- Verify step caused model to re-compute with different logic -> got 80 (WRONG)
- Fix: Added warning to identify specific error before changing answer
- This fix improved val from 90.6% -> 90.8%

## ALL 13 modifications to #21 HURT (range: -1.0% to -2.8%)
- NEVER conclude a prompt is better based on a single train batch.
- Strong evidence of global optimum at #21 for this prompt structure.
\end{lstlisting}

\subsubsection{Terminal-Bench}

\paragraph{Baseline prompt.}
\mbox{}\\\begin{lstlisting}[style=prompt]
You are an expert software engineer and Linux terminal user. You are working inside a Docker container to solve tasks.

For each task, you will receive instructions and can execute bash commands. Output each command in a ```bash code block. You will see the stdout/stderr output after each command.

Work step by step:
1. Read the task carefully
2. Explore the environment (ls, cat, etc.)
3. Implement the solution
4. Verify your work
5. When done, respond with: DONE
\end{lstlisting}

\paragraph{ReASearch optimized prompt.}
\mbox{}\\\begin{lstlisting}[style=prompt]
You are an expert software engineer and Linux terminal user. You are working inside a Docker container to solve tasks.

For each task, you will receive instructions and can execute bash commands. Output each command in a ```bash code block. You will see the stdout/stderr output after each command.

## Strategy

**Before implementing anything:**
1. Explore the environment: `ls -la /app` and read all relevant files
2. Find test/check scripts: `find / -maxdepth 4 -name "test*" -o -name "check*" -o -name "verify*" 2>/dev/null | grep -v proc`
3. Read the test scripts to understand EXACTLY what is being checked

**Implementation:**
4. Implement the solution based on what the tests require
5. Run the tests -- if they fail, read the error and fix the root cause
6. Repeat until tests pass

**Before DONE:**
- Confirm your solution actually works (run tests, check output files exist, verify services are running)
- Only say DONE when you are confident the solution is correct

## Important Rules
- **Install missing tools**: `apt-get install -y <package>` or `pip install <package>`
- **Heredoc safety**: The closing delimiter (EOF, PY, etc.) must be on its OWN line with NOTHING after it
- **Background services**: After starting, verify with `curl http://localhost:PORT` or `ps aux | grep <name>`
- **Long tasks**: If a command might timeout, run it with `timeout 300 <command>` and check exit code

When done, respond with: DONE
\end{lstlisting}

\paragraph{GEPA optimized prompt.}
\mbox{}\\\begin{lstlisting}[style=prompt]
You are an expert software engineer and Linux terminal user working inside a Docker container. You will receive a short task name (e.g., train-fasttext, mcmc-sampling-stan, compile-compcert, crack-7z-hash, build-cython-ext, bn-fit-modify, chess-best-move, git-multibranch). Execute bash commands step by step and print each command in a single ```bash code block; you will see stdout/stderr after each command. When you need to create multi-line files or run inline Python/R code, use proper here-docs with the closing token on its own line and no extra spaces. Do not include stray backticks inside command blocks. Avoid chaining too many operations in one long command to prevent here-doc parsing errors; split into multiple commands. Always verify results and end with DONE.

General workflow for any task:
1) Read the task carefully and identify keywords (e.g., parquet/fastText, Stan/R, OCaml/CompCert, 7z/John, Cython, BN csv, chess PNG/EXIF/Stockfish, git hooks/nginx/sshd).
2) Explore the environment:
   - ls -la; find . -maxdepth 3 -type f -ls
   - Print key files (e.g., head -n 5 file.csv)
   - Check tools (which python3/R/g++/make/git/7z, versions)
3) Implement the solution incrementally.
4) Verify your work (list outputs, show heads, run checks).
5) End with: DONE

Robust command and here-doc guidance:
- Each command must be inside its own ```bash code block and contain only the command text (no extra backticks).
- For inline Python:
  python3 - <<'PY'
  # your Python code
  PY
- For inline R:
  Rscript - <<'RS'
  # your R code
  RS
- For writing files:
  cat > path/to/file <<'EOF'
  file content
  EOF
- Keep the here-doc terminator (PY/RS/EOF) on a line by itself with no trailing spaces.
- If installing packages: run apt-get update first; use DEBIAN_FRONTEND=noninteractive to avoid prompts. Install only what you need.

Task-specific playbooks and domain details:

train-fasttext:
- Data is typically Parquet at:
  - data/train-00000-of-00001.parquet
  - data/test-00000-of-00001.parquet
  with columns: label (int), text (string).
- Convert to fastText supervised format:
  Each line: __label__<label> <cleaned_text>
  Use pyarrow to stream batches and write /tmp/ft_train.txt and /tmp/ft_test.txt.
- Training:
  - Try: python3 -m pip install --no-cache-dir fasttext==0.9.2 (or fasttext-wheel if needed).
  - Train with supervised options (e.g., lr=0.5, epoch=5, wordNgrams=2, dim=100).
  - Save model to /app/ft_model.bin and evaluate on test file using fastText's test function.
- Verify: show first lines of generated txt files; print training logs; print test accuracy.

mcmc-sampling-stan:
- R 4.3.3 is commonly available; g++ is present on Ubuntu base images.
- Prefer cmdstanr for reliable compilation:
  - Rscript to install:
    install.packages(c("cmdstanr","posterior"), repos="https://cloud.r-project.org")
    library(cmdstanr); check_cmdstan_toolchain()
    install_cmdstan(dir="~/.cmdstan", cores=2)
  - Write a simple Stan model to a file via here-doc (e.g., logistic or normal model).
  - Compile and sample with cmdstanr; save draws and summary to /app.
- Alternatively, use cmdstanpy in Python if R install is problematic.
- Verify: show cmdstan/version, sampling summary (R-hat, ESS), and paths to outputs.

compile-compcert:
- Dependencies: build-essential, m4, ocaml, ocaml-findlib, menhir, curl, wget, pkg-config (gcc/clang).
- Steps:
  - apt-get update && apt-get install -y build-essential ocaml ocaml-findlib menhir m4 curl wget pkg-config
  - Download a CompCert release tarball (e.g., https://compcert.org/download/compcert-3.13.1.tgz).
  - Extract; configure for x86_64:
    ./configure x86_64-linux -toolchain gcc
  - Build: make -j"$(nproc)"
  - Verify: run ./ccomp -version
- If gcc is missing, install it. If build fails due to OCaml version, install appropriate OCaml packages.

crack-7z-hash:
- secrets.7z is present in /app; inside, a file "secrets/secret_file.txt" is encrypted (7zAES).
- Use John the Ripper. Often a local repo exists at ./john with a run dir:
  - If john/run/7z2john.py and john/run/john exist, use them.
  - Else: apt-get install -y john p7zip-full.
- Steps:
  - Generate hash: python3 john/run/7z2john.py secrets.7z > /tmp/secrets.hash  (or 7z2john if in PATH)
  - Crack: john/run/john /tmp/secrets.hash  (or john /tmp/secrets.hash)
  - Show cracked password: john/run/john --show /tmp/secrets.hash
  - Extract: 7z x -pPASSWORD ./secrets.7z -o/tmp/secrets
  - Save the plaintext (e.g., secrets/secret_file.txt) to /app/solution.txt and print it.
- Verify extraction succeeded by listing and cat the file.

build-cython-ext:
- Python may be 3.13; third-party packages might not support it. For reliability, build a minimal Cython extension yourself.
- Steps:
  - python3 -m pip install --no-cache-dir -U pip setuptools wheel "Cython>=3.0"
  - Create files:
    - myext/fastsum.pyx (e.g., def fastsum(arr): return sum(arr))
    - setup.py with ext_modules from Cython.Build.cythonize
  - Build: python3 setup.py bdist_wheel
  - Install the wheel: python3 -m pip install dist/*.whl
  - Test importing and running the function.
- Verify by running a small benchmark or unit test and printing results.

bn-fit-modify:
- Input CSV at /app/bn_sample_10k.csv with columns: "U","Y","R","D","M" (floats).
- Learn a DAG with exactly 6 edges based on absolute Pearson correlations between variables:
  - Constraint: U has no parents (any U edge should be oriented U -> other).
  - For other edges, orient deterministically (e.g., alphabetical parent-last: parent is the alphabetically later letter; ensure no cycles).
- Fit linear Gaussian CPDs (OLS):
  - For root nodes: mean and residual variance.
  - For others: intercept, coefficients for parents, residual variance.
- Intervene on Y with do(Y=0.0):
  - Remove incoming edges to Y; in sampling, set Y to exactly 0.0, variance ~0.
- Sample 10,000 rows from the intervened BN in topological order and save to /app/final_bn_sample.csv.
- Also save:
  - /app/learned_dag.csv with header "to,from".
  - /app/intervened_dag.csv with header "to,from" (Y has no parents).
- Verify: head of outputs; basic stats.

chess-best-move:
- chess_board.png may contain a FEN string in metadata (EXIF or textual chunks).
- Steps:
  - apt-get install -y libimage-exiftool-perl; run exiftool chess_board.png and grep for "FEN" or "Position".
  - If FEN found, install stockfish: apt-get install -y stockfish; use python3 with python-chess (pip install chess) to evaluate best move with Stockfish.
  - If no FEN, attempt strings on the PNG to find FEN-like text; as a fallback, clearly report inability to parse board visually (no OCR/vision).
- Verify: print best move and evaluation.

git-multibranch:
- Goal: bare repo with post-receive hook deploying:
  - branch "main" to /var/www/html
  - branch "dev" to /var/www/html/dev
- Steps (split into separate commands to avoid here-doc parsing issues):
  1) Install: apt-get update; apt-get install -y --no-install-recommends openssh-server nginx git openssl ca-certificates
  2) Setup sshd: mkdir -p /var/run/sshd; useradd -m -s /bin/bash git (if missing); echo "git:password" | chpasswd; enable PasswordAuthentication in /etc/ssh/sshd_config; ssh-keygen -A
  3) Create bare repo: mkdir -p /git; chown git:git /git; git init --bare /git/project; chown -R git:git /git/project
  4) Create hook with here-doc:
     cat > /git/project/hooks/post-receive <<'HOOK'
     #!/bin/sh
     set -e
     REPO_DIR=/git/project
     deploy() {
       BR="$1"
       OUT="$2"
       if git --git-dir="$REPO_DIR" show-ref --verify --quiet "refs/heads/$BR"; then
         mkdir -p "$OUT"
         git --git-dir="$REPO_DIR" --work-tree="$OUT" checkout -f "$BR" -- .
         find "$OUT" -type d -print0 | xargs -0 chmod 755 || true
         find "$OUT" -type f -print0 | xargs -0 chmod 644 || true
       fi
     }
     deploy main /var/www/html
     deploy dev /var/www/html/dev
     HOOK
     chmod +x /git/project/hooks/post-receive
  5) Nginx TLS on 8443: generate self-signed cert with openssl and write a server block to /etc/nginx/sites-available/git_deploy.conf via here-doc; symlink into sites-enabled; nginx -t; restart/reload.
  6) Start sshd: /usr/sbin/sshd
- Verify: nginx is running; the hook exists and is executable; print its content; (optional) push sample content to main/dev and verify deployment.

Extra reliability tips:
- Use which and --version to inspect tools.
- When installing Python/R packages, set CRAN repos explicitly and use --no-cache-dir for pip.
- Guard non-critical commands with || true to keep flow going.
- After generating files, always ls -l and head to confirm success.
- Keep outputs in /app unless specified otherwise (e.g., /app/ft_model.bin, /app/solution.txt, /app/final_bn_sample.csv).

When the task steps are complete and verified, print: DONE
\end{lstlisting}

\paragraph{Agent lessons.}
\mbox{}\\\begin{lstlisting}[style=lesson]
## What Worked

### Candidate #5 -- 71.2% val (BEST)
- Candidate #3 + explicit heredoc safety rule
- "The closing delimiter (EOF, PY, etc.) must be on its OWN line with NOTHING after it"
- High variance (std 10.5%): runs were 77.3%, 77.3%, 59.1%

### Candidate #3 -- 63.6% val
- **Key insight**: "Find test scripts FIRST, read them, implement to pass them"
- `find / -maxdepth 4 -name "test*" -o -name "check*" -o -name "verify*" 2>/dev/null | grep -v proc`
- Reading test scripts before implementing tells the model EXACTLY what's needed
- Train batch: 5/6 (83%) on previously-failing examples [0,5,14,16,17]

### What the strategy-first approach fixed (vs baseline):
- [0] asyncio: 0->1 (found test script -> understood exact behavior)
- [5] chess: 0->1 (found test script -> understood output format)
- [6] semantic search: 0->1 (found test script -> understood requirements)
- [13] git hooks: 0->1 (heredoc safety rule)
- [14] CWE: 0->1 (found test script -> understood what to check)
- [15] PyPI server: 0->1 (found test script -> kept server running)
- [16] DNA gblock: 0->1 (found test script -> understood requirements)
- [17] SQL: 0->1 (found test script -> ran and verified)
- [18] 7z cracking: 0->1 on train (john/7z2john hint) -- but didn't improve val

## What Didn't Work

- Candidate #1 (baseline): 34.8% val -- basic step-by-step prompt
- Candidate #2 (rules-heavy): 33.3% val -- adding many rules HURT performance
- Candidate #4 (refined find command): 60.6% val -- changed find syntax, got worse
- Candidate #6 (added john hint): 68.2% val -- slightly worse than #5 despite train improvement

## Failure Patterns Still Present

- [7] R/Stan: compilation timeout (exit 124)
- [8] CompCert: compilation timeout (exit 124)
- [11] OCaml GC fix: too complex, hits turn limit
- [18] 7z password cracking: inconsistent
- [19] fasttext: timeout
- [1] reverse engineering: very hard

## Candidate History

- Candidate #1: 34.8% val -- Baseline
- Candidate #2: 33.3% val -- Rules-heavy (WORSE)
- Candidate #3: 63.6% val -- Strategy-first: find tests, read, implement
- Candidate #4: 60.6% val -- Refined find command (WORSE than #3)
- Candidate #5: 71.2% val -- #3 + heredoc safety rule (BEST)
- Candidate #6: 68.2% val -- #5 + john password hint (slightly worse)

## Key Lessons

1. **Find test scripts first** is the single most impactful change (+28.8% from baseline)
2. **Heredoc safety** (closing delimiter alone on its own line) fixes a common failure mode
3. **Concise prompts** outperform rule-heavy prompts
4. **Don't change what works** -- changing find command syntax hurt
5. **High variance** -- val std is 5-10%, so single runs are unreliable
6. **Specific hints** (john, fasttext) help on train but don't reliably improve val
\end{lstlisting}

\subsection{Program evolution}

\subsubsection{Circle packing (n=29, max-sum)}

\paragraph{Agent lessons.}
\mbox{}\\\begin{lstlisting}[style=lesson]
## What Worked
1. Hexagonal initialization + L-BFGS-B with progressive penalty method -- converges reliably to 2.8369 (score 0.9982)
2. Random perturbations of hex init (sigma = r_init * 0.15, seed=0, 100 trials) -- accidentally found the 2.8427 solution
3. Variable radii are essential; corner circles get larger radii (up to 0.1228), gap circles get smaller (down to 0.0666)
4. Vectorized gradient (fully numpy, no Python loops) gives 7x speedup, ~5x speedup in full optimization

## Three Known Local Optima
1. 2.84267 (global optimum, score=1.0002) -- our hardcoded solution
2. 2.83688 (second-best) -- found by many random seeds
3. 2.83418 (third-best) -- found occasionally

## Solution Structure
The 2.8427 solution has a 5x6 arrangement (5 circles per row, 6 rows):
- 70 active contacts (hyperstatic, very rigid)
- 4 "small" circles with r~0.067 in awkward positions
- Corner circles have larger radii (r~0.11-0.12)

## Key Insight
The solution 2.842669 is very likely the global optimum:
- All optimization methods converge to it
- It's hyperstatic (70 contacts > 57 isostatic threshold)
- LP gradient: all directions give negative gradient (confirms local max)
- Second derivative of LP objective: strongly negative (-19949) -> strongly concave locally
- AlphaEvolve target is 2.842, our solution is 2.84267 (score=1.0002) -- we BEAT the target

## Exhaustive Search (all returned to 2.84267 or worse)
Basin hopping, SA with LP, differential evolution, Nelder-Mead, augmented Lagrangian, SLSQP, CMA-ES, contact graph perturbations, genetic algorithms, coordinate descent, COBYLA, topology perturbation, circle removal/reinsertion, force-directed layout, dual annealing, 1000+ random seeds across 15+ initialization strategies.
\end{lstlisting}

\subsubsection{Heilbronn triangle (n=12)}

\paragraph{Agent lessons.}
\mbox{}\\\begin{lstlisting}[style=lesson]
## Key Breakthrough: D4 Symmetry
- The optimal Heilbronn configuration for N=12 has D4 symmetry (8-fold: 4 rotations + 4 reflections)
- This reduces the problem from 24 variables to just 2 parameters: a (boundary x-coord) and b (interior y-coord)
- Constraint: (b-0.5)^2 = a(1-a)
- The exact optimum is the real root of the cubic: 2v^3 - v - 0.5 = 0
  where v = 1-a = 0.884646177119316...
- Min area = 0.032598858692 (ratio = 0.999965 to known best)
- 20 minimum triangles all have the same area (characteristic of global optimum)
- Verified with mpmath 50-digit precision

## What Worked
- 4-fold rotational symmetry reduced 24-variable problem to 6 variables
- D4 symmetry further reduced to just 2 parameters
- Multi-stage L-BFGS-B with log-sum-exp smooth surrogate
- Halton sequence initialization achieved 0.9006 before discovering symmetry

## What Didn't Work
Simple SA, CMA-ES from random starts, 2-fold symmetry (only 0.935), diagonal reflection (only 0.85), circle-based init (only 0.85), basin hopping, 3-fold symmetry (only 0.68), D6 symmetry (only 0.55), evolutionary strategy, differential evolution.

## Resolution
For n=12, return the hardcoded D4 config immediately (<1ms) with score 1.0000. No optimization can improve the D4 config -- it is the global optimum.
\end{lstlisting}

\subsubsection{EPLB}

\paragraph{Agent lessons.}
\mbox{}\\\begin{lstlisting}[style=lesson]
## What Worked
- Switching from hierarchical replication to a flat, global strategy significantly improved expert balancedness.
- Greedy global replication with slight concavity (gamma ~ 0.825) plus local refinement yields the best combined score by reducing over-concentration while still favoring heavy experts.
- A moderate local refinement budget (~48-52 iterations) strikes a good balance between balancedness and runtime.
- Deterministic, robust mapping: dynamic maxlogcnt sizing and clamped denominators avoid indexing and division errors.

## What Didn't Work
- Temporal smoothing of weights (alpha/beta) degraded balancedness.
- Hamilton/largest-remainder apportionment was brittle and underperformed greedy+refinement.
- Water-filling base allocation increased complexity and lagged in quality and speed.
- Enforcing at least one replica per expert caused infeasible allocations.
- Over-tuning concavity (too high/low gamma) worsened balance; sweet spot is tightly around ~0.825.
\end{lstlisting}

\subsubsection{Transaction scheduling}

\paragraph{Agent lessons.}
\mbox{}\\\begin{lstlisting}[style=lesson]
## What Worked
- A very lean pipeline: (1) lightweight conflict-graph pre-signal (directional penalties + s-scores) to guide node order, (2) insertion-by-true-cost for each txn, (3) only adjacent-swap polishing.
- Ensuring sequence validity even when time budget is exceeded (append remaining txns, fix permutation) prevented zero-score runs.
- Adding lightweight targeted reinsertion of high-conflict transactions between construction and adjacent swaps significantly improved makespan (combined makespan down to ~238).
- True-cost insertion is key -- relying solely on graph heuristics without verifying via true eval_cost yielded worse quality.

## What Didn't Work
- Overly complex construction or heavy local operators (beam, ruin, simulated annealing, segment reversal) consumed time with little gain.
- Larger reinsertion or broad windowed local searches consumed time with little gain compared to compact reinsertion.

## Key Insight
The best balance emphasizes true-cost-guided insertion plus limited adjacent-swap local search; complex global search structures did not pay off under the 16s budget.
\end{lstlisting}

\subsubsection{ARC-AGI-2}


\paragraph{Agent lessons.}
\mbox{}\\\begin{lstlisting}[style=lesson]
# Let me think about what variation to try for attempt_2
# The key insight: attempt_1 uses centroid tie-breaking -> center=18.5 for color3
# attempt_2 uses integer preference -> center=18.0 for color3
#
# Both are correct on training. Together they cover both tie-breaking cases.
#
# One more thing I haven't tried: what if attempt_2 uses a "weighted centroid"
# where cells that already have mirrors get more weight?
# Or: what if attempt_2 tries to maximize the number of PAIRED cells (not minimize removals)?
# These are equivalent, so same result.
#
# Let me try: what if attempt_2 uses the OUTPUT shape's bounding box center
# as a secondary criterion? After removing cells, the output bounding box center
# should equal the symmetry axis.
# For center=18.0: output cols are 17,18,19 -> bbox center = 18.0  (matches axis)
# For center=18.5: output cols are 17,18,19,20 -> bbox center = 18.5  (matches axis)
# Both satisfy this criterion, so it doesn't help.
#
# I think the current solution is optimal. Let me just submit the current best.
print("Current solution is optimal. Submitting.")
\end{lstlisting}

\subsection{ML workflow optimization}

\subsubsection{NanoGPT}

\paragraph{Agent lessons.}
\mbox{}\\\begin{lstlisting}[style=lesson]
## Current Best Config (val_bpb=0.969591)
DEPTH=9, ASPECT_RATIO=57 (dim=640, 9 layers, 80.9M params), TOTAL_BATCH_SIZE=2^17, DEVICE_BATCH_SIZE=64 (~2720 steps), WARMDOWN_RATIO=0.66, UNEMBEDDING_LR=0.004, WEIGHT_DECAY=0.07, MATRIX_LR=0.036, short_window=128, WINDOW_PATTERN=SSSSL, SCALAR_LR=0.18, EMBEDDING_LR=0.6, ADAM_BETAS=(0.8, 0.95), FINAL_LR_FRAC=0.02, Muon momentum ramp over 500 steps.

## Key Insights (31 total)
1. More steps > larger model for this compute budget
2. Optimal batch size: 2^17 (131K tokens/step) with DEVICE_BATCH_SIZE=64 -> ~2720 steps
3. Optimal model: DEPTH=9, dim=640, ~80.9M params
4. relu^2 > SwiGLU for this setup
5. SSSSL window pattern is better than SSSL (4 short, 1 long)
6. short_window=128 is optimal (long_window // 16)
7. Linear LR warmdown is better than cosine
8. Decaying weight decay (linear to 0) is better than constant
9. HEAD_DIM=128 is better than 64
10. MQA (n_kv_head=1) is much worse
11. No gradient clipping is better
12. dim=640 > dim=512 > dim=768 -- 640 is the sweet spot
13. DEPTH=9 > DEPTH=10 > DEPTH=8 at dim=640
14. Lower WEIGHT_DECAY is better with more steps (0.07 vs 0.14)
15. FINAL_LR_FRAC=0.02 is better than 0.0 (tiny final LR helps)
16. Muon momentum ramp over 500 steps is better than 300 steps

## Optimization Trajectory
0.992894 -> 0.984637 (batch size) -> 0.981951 (depth/dim) -> 0.977265 (SSSSL) -> 0.974632 (dim=640) -> 0.973344 (batch 2^17) -> 0.972636 (matrix_lr) -> 0.971025 (wd) -> 0.970577 (final_lr) -> 0.969591 (muon ramp)
\end{lstlisting}

\subsubsection{IMG-100}

\paragraph{Agent lessons.}
\mbox{}\\\begin{lstlisting}[style=lesson]
## What Worked (64.28% -> 81.12%)
- ResNet-18 + CutMix + label smoothing 0.1 + nesterov SGD + warmup cosine LR -> 64.28%
- OneCycleLR (max_lr=0.2, 45 epochs) -> 78.16% (HUGE improvement)
- TTA with horizontal flip -> 79.2%
- ResNet [2,3,3,2] blocks -> 79.46%
- PreActResNet (pre-activation) -> 80.04%
- Wider channels 64->192->384->768 -> 80.51%
- WD=4e-4, LS=0.12 -> 80.95%
- CutMix(60%)+Mixup(35%) -> 81.12% (CURRENT BEST)

## Key Insights
- With 5 min budget, ResNet-18 gets ~40 epochs. Bigger/wider models get fewer epochs and do worse.
- OneCycleLR is the key: complete the full schedule for best results.
- Heavy augmentation (TrivialAugment, AutoAugment) needs many epochs to converge -- bad for 5 min.
- EMA has a bug (BN running stats issue) -- avoid EMA.
- AMP mixed precision hurts accuracy; torch.compile has too much overhead.
- max_lr=0.2 is the sweet spot for OneCycleLR.
- Vertical flip TTA is bad for CIFAR (objects don't appear upside down).
- After 7 pitfall bullets, adding ANY more consistently hurts.
- "Pitfall Saturation" also applies to augmentation: CutMix(60%)+Mixup(35%) is the ceiling; CutMix(60%)+Mixup(40%) is worse.
- ~70 experiments documented with precise ablations of every hyperparameter.
\end{lstlisting}

\subsubsection{Atari (Q*bert)}

\paragraph{Agent lessons.}
\mbox{}\\\begin{lstlisting}[style=lesson]
## What Worked (475 -> 5000 reward)
- 16 envs, 256 steps, lr=3e-4 with linear annealing, entropy=0.02, 8 minibatches: 775
- Wider network (1024 hidden) + GAE_lambda=0.97: 925
- Deeper FC (1024->512->actions) + max_grad_norm=1.0: 1275
- NUM_STEPS=512, 16 minibatches: 1325
- 32 envs, 32 minibatches, 512 steps: 3875 -- HUGE JUMP! More envs = much better!
- IMPALA-style ResNet CNN: 4650 -- ResNet CNN is much better than Nature DQN CNN!
- lr=2.5e-4, GAE=0.95, entropy=0.01: 5000

## Key Insights
- CRITICAL: IMPALA ResNet CNN is much better than Nature DQN CNN.
- CRITICAL: 32 envs with 512 steps and 32 minibatches is optimal.
- Results are VERY NOISY -- same config can give very different results.
- 32 envs is optimal (16 much worse, 40 slightly worse, 48 much worse, 64 terrible).
- 512 steps is optimal (256 slightly worse, 1024 much worse).
- 4 PPO epochs is optimal (3 and 5 are worse).
- entropy=0.01 is the sweet spot for IMPALA CNN.
- Linear LR annealing better than cosine.
\end{lstlisting}

\subsubsection{MuJoCo (HalfCheetah)}

\paragraph{Agent lessons.}
\mbox{}\\\begin{lstlisting}[style=lesson]
## What Worked (3512 -> 7622 reward)
- Larger network (512 hidden): 3512 -> 5000
- LayerNorm in all networks: stabilizes training
- UTD=2 (2 gradient updates per step) + MIN_REPLAY_SIZE=1000 + BATCH_SIZE=512
- LR=1e-3 + TD3-style delayed policy updates (freq=2): -> 5525
- Observation normalization (running mean/std): -> 5870
- Huber loss for Q-function: -> 6242
- INIT_ALPHA=0.5 + LOG_STD_MAX=4: HUGE improvement -> 7622!
  (More exploration initially + wider policy distribution)

## Key Observations
- INIT_ALPHA=0.5 + LOG_STD_MAX=4 was a huge win: the system benefits from MORE exploration, not less.
- LR=1e-3 is critical -- lower LR leads to catastrophic failure.
- HIDDEN_DIM=512 is the sweet spot -- 1024 is too slow, 384 is too small.
- GAMMA=0.99 is critical -- 0.995, 0.985, 0.98 all hurt.
- Many changes cause catastrophic failures (~1700-1900) -- the config is very sensitive.
- Gradient clipping, reward normalization, reward clipping, N-step returns all hurt.
- SiLU, ELU activations are worse than ReLU.
- 3-layer networks are worse than 2-layer.
- TD3 is close (7487) but slightly worse than SAC (7622).
- ~69 failed experiments documented.
\end{lstlisting}

\subsubsection{Market prediction}

\paragraph{Agent lessons.}
\mbox{}\\\begin{lstlisting}[style=lesson]
## What Worked (in order of discovery, up to val_corr=0.109111)
- RIDGE_ALPHA=65000.0: val_corr=0.093776 (sweet spot for Ridge alone)
- MLP improvements (Adam, [64,32], dropout=0.35, weight_decay=1e-3)
- APPROACH2_WEIGHT=0.20: val_corr=0.098713
- Gaussian input noise (std=0.1): -> 0.099984
- MLP_LR=0.0005 (was 0.001): -> 0.100460
- MSE_WEIGHT=0.5, CORR_WEIGHT=0.5 (was 0.8/0.2): -> 0.101154
- Huber loss (delta=0.5) instead of MSE: -> 0.101851
- AdamW instead of Adam: -> 0.103191
- LR warmup (7 epochs) + cosine decay: -> 0.105199
- MLP_EPOCHS=40, patience=20: -> 0.107698
- RETRAIN_VAL_RATIO=0.15: -> 0.109111

## Key Insights
- Disable AutoEncoder: much worse (0.081) -- AE is critical!
- Every hyperparameter has a sharp sweet spot (e.g., AE_LATENT_DIM=8; not 4 or 16).
- CORR_CLUSTER_THRESHOLD=0.5: catastrophic (0.028).
- EMA (decay=0.995): much worse (0.088).
- Feature standardization for Ridge: much worse (0.098).
- ~45 failed experiments documented with precise ablations.
\end{lstlisting}

\end{document}